\documentclass[10pt,twocolumn,letterpaper]{article}

\usepackage[pagenumbers]{cvpr} 

\newcommand{\xmark}{\ding{55}}%

\usepackage{bm}
\usepackage{amsmath}
\usepackage{graphicx}
\usepackage{multirow}
\usepackage{makecell}
\usepackage{colortbl}
\usepackage{pifont}

\usepackage{algorithm}      
\usepackage{algpseudocode}  
\usepackage{amsmath}        
\usepackage{amssymb}      

\usepackage{multicol}
\usepackage{longtable}
\usepackage{tcolorbox}

\usepackage{subcaption}

\usepackage{pifont}

\definecolor{darkpastelred}{rgb}{0.76, 0.23, 0.13}

\newcommand{\cmark}{\ding{51}}%
\newcommand{\magma}{\texttt{Magma}\xspace}

\newcommand\blfootnote[1]{\begingroup\renewcommand\thefootnote{}\footnote{#1}\addtocounter{footnote}{-1}\endgroup}

\definecolor{commentcolor}{RGB}{34,139,34}
\definecolor{cvprblue}{rgb}{0.21,0.49,0.74}
\usepackage[pagebackref,breaklinks,colorlinks,allcolors=cvprblue]{hyperref}

\def\paperID{1545} 
\def\confName{CVPR}
\def\confYear{2025}

\title{\includegraphics[height=0.8cm]{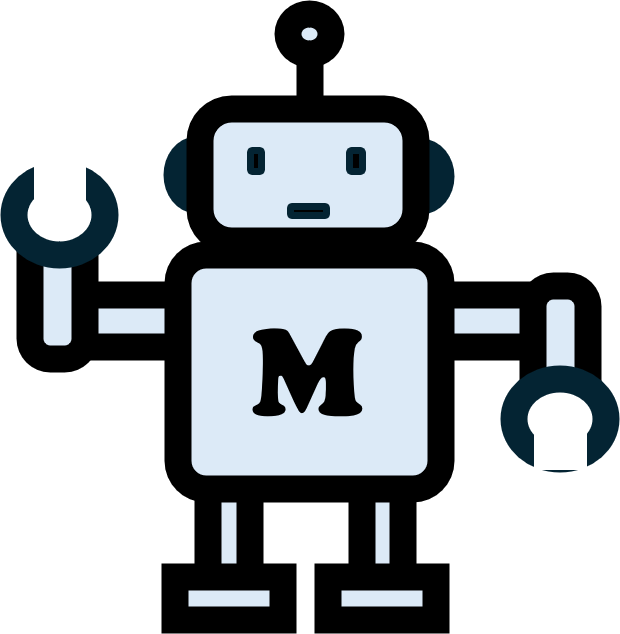} Magma: A Foundation Model for Multimodal AI Agents}

\author{Jianwei Yang\textsuperscript{1*$\dagger$} \ \ \ Reuben Tan\textsuperscript{1$\dagger$} \ \ \ Qianhui Wu\textsuperscript{1$\dagger$} 
\ \ \ Ruijie Zheng\textsuperscript{2$\ddagger$} \ \ \  Baolin Peng\textsuperscript{2$\ddagger$} \ \ \ Yongyuan Liang\textsuperscript{2$\ddagger$} \\
Yu Gu\textsuperscript{1} \ \ \ Mu Cai\textsuperscript{3} \ \ \ Seonghyeon Ye\textsuperscript{4} \ \ \ Joel Jang\textsuperscript{5} \ \ \ Yuquan Deng\textsuperscript{5} \ \ \ Lars Liden\textsuperscript{1} \ \ \ Jianfeng Gao\textsuperscript{1$\bigtriangledown$} \\
$^{1}$\small{Microsoft Research}, $^{2}$University of Maryland, $^{3}$University of Wisconsin-Madison
\\
$^{4}$\small{KAIST}, $^{5}$University of Washington \\
\url{https://microsoft.github.io/Magma}
\\
}

\begin{document}
\twocolumn[{%
\renewcommand\twocolumn[1][]{#1}%
\maketitle
\begin{center}
   \captionsetup{type=figure}
    \vspace{-0.8cm}    
\includegraphics[width=1.0\linewidth]{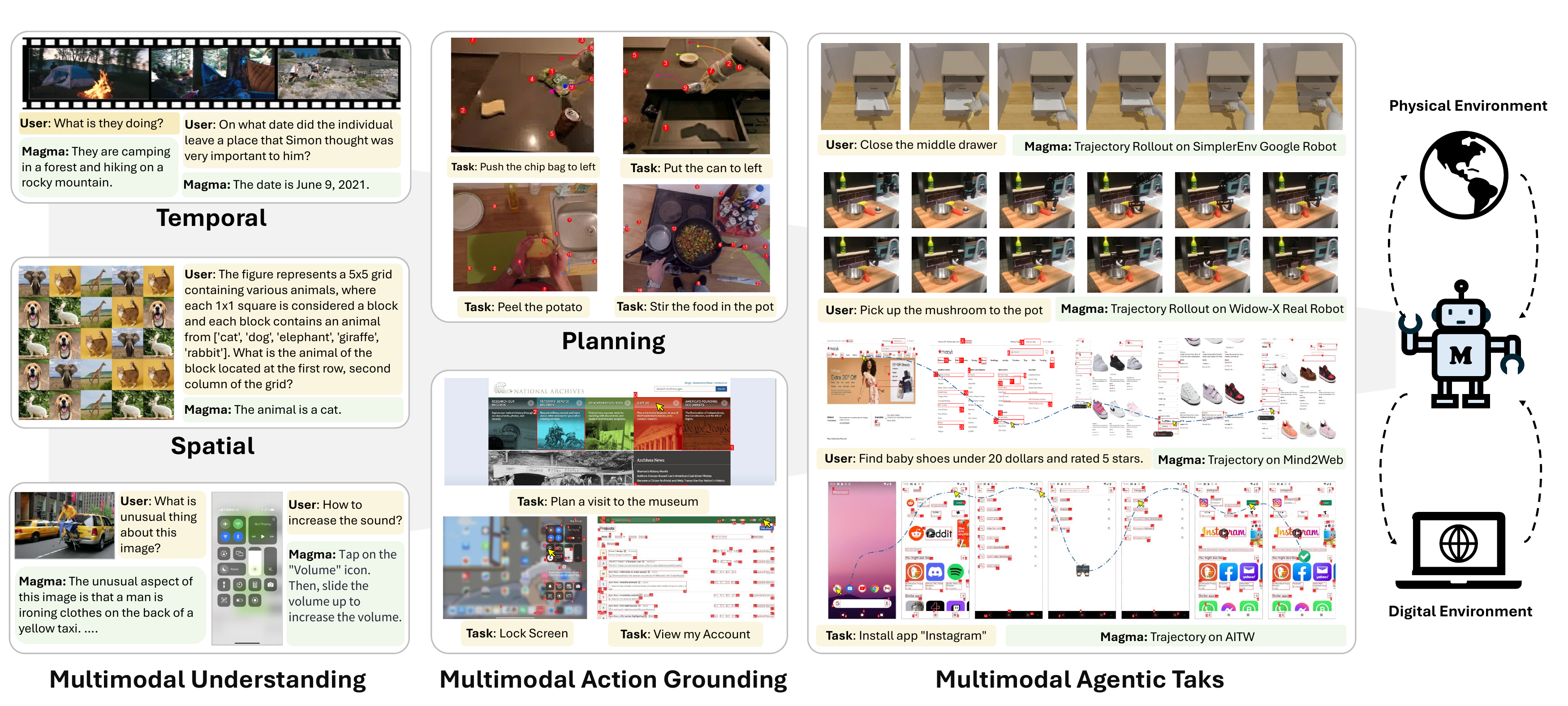}    
    \vspace{-0.5cm}
    \caption{We introduce \magma, the \textit{first} foundation model that is capable of interpreting and grounding multimodal inputs within its environment. Given a described goal, \magma is able to formulate plans and execute actions to achieve it. By effectively transferring knowledge from freely available visual and language data, \magma bridges verbal and spatial intelligence to navigate complex tasks.}
    \label{fig:teaser} 
\end{center}   
}]

\blfootnote{$^{\dagger}$ First Authors; $^{\ddagger}$ Second Authors; $^*$ Project Lead; $^{\bigtriangledown}$ Leadership}


\begin{abstract}
We present \magma, a foundation model that serves multimodal AI agentic tasks in both the digital and physical worlds.  \magma is a significant extension of vision-language (VL) models in that it not only retains the VL understanding ability (verbal intelligence) of the latter, but is also equipped with the ability to plan and act in the visual-spatial world (spatial-temporal intelligence) and complete agentic tasks ranging from UI navigation to robot manipulation. To endow the agentic capabilities, Magma is pretrained on large amounts of heterogeneous datasets spanning from images, videos to robotics data, where the actionable visual objects (e.g., clickable buttons in GUI) in images are labeled by Set-of-Mark (SoM) for action grounding, and the object movements (e.g., the trace of human hands or robotic arms) in videos are labeled by Trace-of-Mark (ToM) for action planning.  
Extensive experiments show that SoM and ToM reach great synergy and facilitate the acquisition of spatial-temporal intelligence for our \magma model, which is fundamental to a wide range of tasks as shown in Fig.~\ref{fig:teaser}. In particular, \magma creates new state-of-the-art results on UI navigation and robotic manipulation tasks, outperforming previous models that are specifically tailored to these tasks. On image and video-related multimodal tasks, Magma also compares favorably to popular large multimodal models that are trained on much larger datasets. We make our model and code public for reproducibility\footnote{\url{https://microsoft.github.io/Magma}}.

\end{abstract}
\section{Introduction}
\label{sec:intro}
A long-standing research topic of AI is to develop autonomous agents that can perceive visual stimuli, language inputs, and other environmentally-grounded data and produce meaningful embodied actions in physical and digital environments to complete specific tasks.

Recently, there has been a growing interest in developing AI agents based on Vision-Language-Action (VLA) models~\cite{kim2024openvla,driess2023palm, brohan2022rt,brohan2023rt,hong2023cogagent,seeclick}. These models are typically pretrained on large amounts of vision-language datasets and then action trajectories to attain ability to take actions given VL inputs. However, due to the inherent difference between various environments~(\eg, 2D digital world and 3D physical ones), VLA models are typically trained separately for simplicity and then used for different tasks. Exemplary models in the digital world include Pix2ACT~\cite{shaw2023pixels}, WebGUM~\cite{furuta2023multimodal}, and Ferret-UI~\cite{you2023ferret} for UI navitation. VLA models in the 3D physical world include RT-2~\cite{brohan2022rt} and OpenVLA~\cite{kim2024openvla} for robotics manipulation.
Although claimed as generalist, most of these models prioritize learning a task-specific action policy at the cost of a significant decline in generic multimodal understanding capabilities, rendering limited genralizability across tasks and domains. 

\begin{figure}
    \centering
    \includegraphics[width=1.0\linewidth]{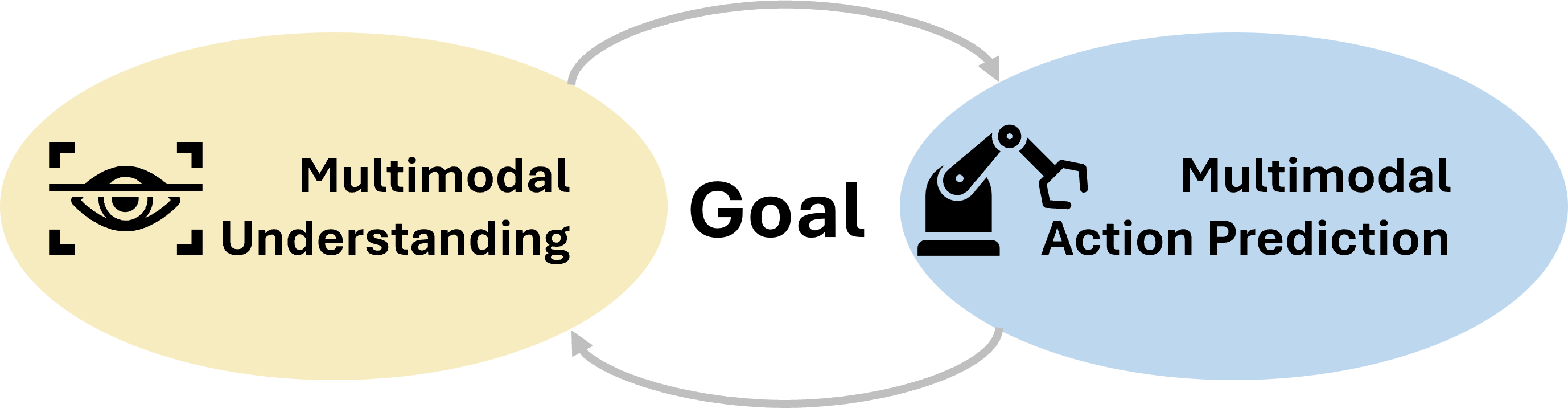}
    \caption{A multimodal AI agent should be capable of mutimodal understanding and action-prediction towards a given goal.}
    \label{fig:intro}
    \vspace{-0.5cm}
\end{figure}

In this research, we strive to develop a foundation model for multimodal AI agents and argue that it requires simultaneously possessing the following capabilities:
\begin{itemize}
    \item \textbf{Multimodal Understanding} to understand multimodal input from various domains (both digital and physical) not only semantically, but also spatially and temporally.
    \item \textbf{Multimodal Action Prediction} to break down the long-horizon task into an accurate action sequence, which can be effectively executed by AI agent systems.
\end{itemize}
Such an agent system should be driven by external goals specified by human commands as shown in Fig.~\ref{fig:intro}.

To endow the broad capabilities, we effectively leverage large amounts of heterogeneous vision-language and action datasets, including UI datasets such as SeekClick~\cite{seeclick}, robotic manipulation dataset OXE~\cite{open_x_embodiment_rt_x_2023}, human instructional videos like Ego-4d~\cite{grauman2022ego4dworld3000hours} and image-text pairs used in LMMs~\cite{liu2023llava,chen2023sharegpt4v}. Instead of sequentially training on one domain and adapting to another, we train a \textit{single} foundation model which can be applied in a zero-shot manner to different downstream tasks in various settings.

Simply combining those datasets, however, does not bring benefits to the foundation model, due to the significant gap between multimodal understanding which is mostly verbal (\ie, textual descriptions for images and videos) and the action-taking tasks which are mostly spatial~(\ie, 2D coordinates for UI or 7-DoF for robot arm). To bridge the gap, we propose two surrogate tasks for model training, action grounding and action planning, by asking the model to predict the proximal action outputs given the visual-spatial observations, represented as images or video frames. 
Specifically, in each image, we label the visual objects that are actionable by \textbf{Set-of-Mark (SoM)} (e.g., clickable buttons in Fig.~\ref{fig:teaser} bottom-middle) and labeled in each video the object movements, which are the results of actions, with \textbf{Trace-of-Mark (ToM)} (e.g., the trace of human hand or robotic arm in Fig.~\ref{fig:teaser} top-middle). In this way, the image and video datasets, which are not labeled with actions, are transformed into ``vision-language-action'' data to morph the gap among different types of tasks. 
We show through extensive empirical studies that SoM and ToM achieve are environment-agnostic and easy to generalize to new agentic tasks, offering an effective and efficient approach to scaling up our \magma model pretraining using large amounts of unlabeled videos, such as raw instructional videos.

To the best of our knowledge, \magma is the first foundation model for multimodal AI agents that can understand multimodal inputs~(see Fig.~\ref{fig:teaser}~left), perform action grounding and planning for the future~(see Fig.~\ref{fig:teaser}~middle), and finally adapt to downstream (unseen) agentic tasks in both the digital and physical environments(see Fig.~\ref{fig:teaser}~right). We evaluated Magma on three task categories: UI navigation (e.g., Mind2Web, AITW), where it has to reason and act in evolving digital environments; vision-language understanding (e.g., GQA, VideoMME), where it grounds language in visual objects and events; and finally robotic manipulation (e.g., Bridge, LIBERO), which tests its 3D spatial intelligence for physical interaction. \magma achieves new SOTA results on UI navigation and robotic manipulation tasks, outperforming even domain-specific models while maintaining strong performance on VL tasks which are comparable to SOTA LMMs. 

In summary, the main contributions of this work are:
\begin{itemize}
    \item We propose \magma, the first foundation model that acquires not only multimodal understanding but also spatial-temporal reasoning abilities for agentic tasks in both digial and physical environments. 
    \item We propose the use of Set-of-Mark and Trace-of-Mark techniques to significantly enhance the spatial-temporal intelligence for action grounding and planning, and allow \magma to be pretrained effectively on large amounts of heterogeneous datasets. 
    \item We curate a large-scale pretraining dataset, which consists of not only open-source VL datasets, but also UI, robotics data and human instructional videos, auto-labeled using SoM and ToM. In total, our training corpus contains approximately 39 million diverse samples.
    \item We extensively evaluate the pretrained \magma model to demonstrate the superior model performance across a wide range of tasks. \magma with a single suite of parameters achieves new SOTA on both robotic manipulation and UI navigation over open-sourced counterparts.    
    \item We show that the proposed \magma pretraining method significantly improves model's verbal and spatial-temporal intelligence abilities. For instance, \magma can achieve SOTA performance on the BLINK dataset without instruction fine-tuning, and SOTA performance on video question-answering benchmarks despite being pretrained on much fewer frames. 
    
\end{itemize}
\section{Related Work}
\label{sec:related}

\textbf{Large Multimodal Models~(LMMs).} Large Language Models (LLMs) like ChatGPT~\citep{chatgpt}, GPT-4~\citep{gpt4}, and Llama~\citep{touvron2023llama} have demonstrated impressive reasoning and generalization capabilities for text. The introduction of models that integrate visual data has brought about a significant shift in the landscape of LLMs, such as GPT-4V(ision)\citep{GPT4V_System_Card}. Building upon open-source LLMs \citep{touvron2023llama,vicuna2023}, a wide range of multimodal models have achieved remarkable progress, led by pioneering models such as LLaVA~\citep{liu2023llava, liu2023improvedllava} and MiniGPT-4~\citep{zhu2023minigpt}, which combine LLMs' capabilities with a CLIP~\citep{radford2021learning} based image encoder. Recently,  a growing number of LMMs have been developed to handle a wider range of tasks and modalities, such as region-level LMMs~\citep{cai2024vipllava, zhang2023gpt4roi, chen2023shikra, peng2023kosmos,zhang2023llavagrounding}, and video LMMs~\citep{lin2023video, zhang2023video, zhang2024llavanextvideo,tan2024koala}. In parallel, more sophisticated benchmarks are proposed to assess these capabilities~\cite{fu2024videommefirstevercomprehensiveevaluation,fu2024blink,cai2024temporalbenchbenchmarkingfinegrainedtemporal}. 

\noindent\textbf{UI Agent in Digital World.} Recently there has been a lot of work on designing autonomous GUI agents to perform tasks in place of human users. One line of work is to train an end-to-end model to directly predict the next action, representative works include Pixel2Act~\cite{pixel2act} and WebGUM\cite{webgum} in web domain, Ferret~\cite{ferretui}, CogAgent~\cite{cogagent}, and Fuyu~\cite{fuyu_8b} in Mobile domain. Another line of work involves leveraging existing multimodal models such as GPT-4V to perform user tasks. Representative works include MindAct~\cite{mind2web}, SeeAct~\cite{zheng2024gpt4vision} in web domain and others ~\cite{gpt4v_wonderland, mobile_agent, aitw} for mobile domain. 
These works often leverage the DOM information in web browsers, or the view hierarchies in mobile apps to get the ground truth position of interactable elements of the screen, and use Set-of-Mark~\cite{setofmark} or more advanced localization model~\cite{lu2024omniparser} to overlay the bounding boxes on top of the screenshot that feed into the vision-language models. 

\noindent\textbf{Vision-Language-Action for Robotics.} Several studies have investigated the application of LMMs in robotics~\citep{brohan2023rt,niu2024llarva,zhu2024vision,li2024llara,kim2024openvla,zheng2024tracevla,ye2024latent}. 
Among these, RT-2~\citep{brohan2023rt} finetuned LMMs on robotic trajectory data, enabling the output of discretized robot action tokens. OpenVLA~\citep{kim2024openvla} is the first open-source VLA foundation that is fine-tuned an open-source Prismatic VLM backbone~\citep{karamcheti2024prismatic}.
LLARVA~\citep{niu2024llarva} generated 2D visual traces for robot arms along with textual representations of actions, using visual trace prediction as an auxiliary task, while TraceVLA~\cite{zheng2024tracevla} used visual trace prompting to improve spatial-temporal awareness of robot policy. Most recently, learning from videos by predicting the latent VQVAE tokens is explored in \cite{cheang2024gr2generativevideolanguageactionmodel,ye2024latentactionpretrainingvideos}. In this work, we follow a similar approach as OpenVLA to represent the action but leverage rich multimodal data far beyond robotics datasets. Also, instead of asking model to predict latent tokens, we propose SoM and ToM techniques to significantly enhance the spatial-temporal intelligence, demonstrating significantly stronger performance and generalization capability for agentic tasks.
\section{Multimodal Agentic Modeling}
\label{sec:formatting}

\subsection{Problem Definition}

A generalist multimodal AI agent should be performant for both multimodal understanding and action-taking. We define a multimodal AI agent ${\pi}$, which takes past visual observations $\mathcal{I} = \{I_1,...,I_k\}$ and a task description $\texttt{task}$ in text as input and outputs a set of $T\geq 1$ tokens $\mathcal{O}$ as:
\begin{equation}
    \mathcal{O} = \mathcal{\pi}(\mathcal{I}, \texttt{task}, \texttt{ctx}) = \{o^{l}_{1},\cdot\cdot\cdot,o^{l}_{T}\}
    \label{equ1:original target}
\end{equation}
where $\texttt{ctx}$ denotes the context, $l \in \{\texttt{verbal}, \texttt{spatial}\}$ indicates if the $i$-th token $o_i$ is a verbal or spatial token. This formula generalizes across different tasks:
\begin{itemize}
    \item \textbf{UI navigation in 2D screenshots}. The task could be ``book a hotel'' and output should include both language tokens denoting the semantic type of action (\eg, ``type'', ``click'', \etc) and the location $(x,y)$ or box $(x,y,w,h)$ to which actions are applied.
    \item \textbf{Robotic manipulation in the 3D world}. For a task like ``close the drawer'', the output consists of 6-DoF displacements $(x,y,z,yaw,pitch,roll)$ of the end effector and, in some cases, one additional dimension to indicate whether the gripper is open or not.
    \item \textbf{Multimodal understanding tasks}. When the task is purely about $\mathcal{I}$, \eg, a VQA task, the problem is reduced to a multimodal understanding task that generates a textual description and/or location of objects for input images/videos.
\end{itemize}

For these seemingly different output modalities, we follow a common practice to transform all output into textual tokens to facilitate model learning. Specifically, we convert 2D actions into a textual dictionary as in~\cite{seeclick}, and represent robot actions with the last 256 discrete language tokens that is barely used in LLMs, following~\cite{kim2024openvla}. Despite such unification into language space, we notice considerable conflicts among tasks, as we will show in our experiments. In what follows, we will discuss how to mitigate such challenges to train agentic foundation on a wide range of datasets.

\subsection{Method}

We approach two key challenges while building a highly capable foundation for the multimodal AI agent. 

\noindent \textbf{Pretraining objectives}: How to build a unified pretraining interface to facilitate joint training? A straightforward way would be to predict the \textit{2D} coordinates for the navigation of the UI, \textit{ 3D} positions for the end effectors, and regular textual outputs for VL tasks. However, in our experiments, we observed that these tasks have inherent domain gaps in both input and output. The former results in a huge search space at the pixel level, and the latter directly predicts the output of proprioceptive action, which is not grounded on the observations of the image. \textit{Can we come up with a surrogate task that can bridge the gap among all tasks?}

\noindent \textbf{Data scaling-up}: Existing vision-language-action data have limited amount and diversity, unlike language or image-text corpus for LLMs and LMMs, respectively. 
For example, the largest open source robotic dataset OXE~\cite{open_x_embodiment_rt_x_2023} consists of around 1M trajectories taken from 22 environments. On the other hand, large-scale image-text datasets like LAION~\cite{schuhmann2021laion} barely contain useful supervisions for action pretraining as they are all static without the notion of action. Videos, however, depict numerous human actions and human-object interactions.\textit{ Can we largely take advantage of these video data for our agentic pretraining?}

In this work, we propose a simple yet effective method to address the aforementioned challenges. Inspired by the generality of Set-of-Mark~(SoM) rompting~\cite{yang2023set}, we employ it to enable the action grounding onto images for both UI and robotic tasks in that model faces much less difficulties to predict the numeric marks for both clickable buttons or robot arms in the image space. We further extend it along the temporal axis and ask the model to predict Trace-of-Mark~(ToM), which forces the model to learn a longer horizon by predicting distant future ``actions'', and more importantly provides an effective way to leverage unlabeled video data. The combination of SoM and ToM enables a seamless synergy across agentic tasks in digital and physical domains, as well as a scalable way to curate ``action'' supervisions from raw videos. We describe them in detail below in Sec.~\ref{sec:som} and~\ref{sec:tom}, respectively. \looseness=-1

\subsubsection{Set-of-Mark for Action Grounding}
\label{sec:som}
SoM prompting~\cite{yang2023set} was first proposed to enhance the grounding capability of GPT-4V and has then been widely adopted for various agentic tasks~\cite{liu2024moka,nasiriany2024pivot,yan2023gpt,huang2024copa,cheng2024spatialrgpt}. Unlike previous works that exploited it for prompting off-the-shelf LMMs to enhance visual-language grounding, here we propose \textit{to train} an agentic model for action grounding, \ie, locating actionable points / regions for a specific task and further predict atomic actions if needed.

\begin{figure}
    \centering
    \includegraphics[width=1.0\linewidth]{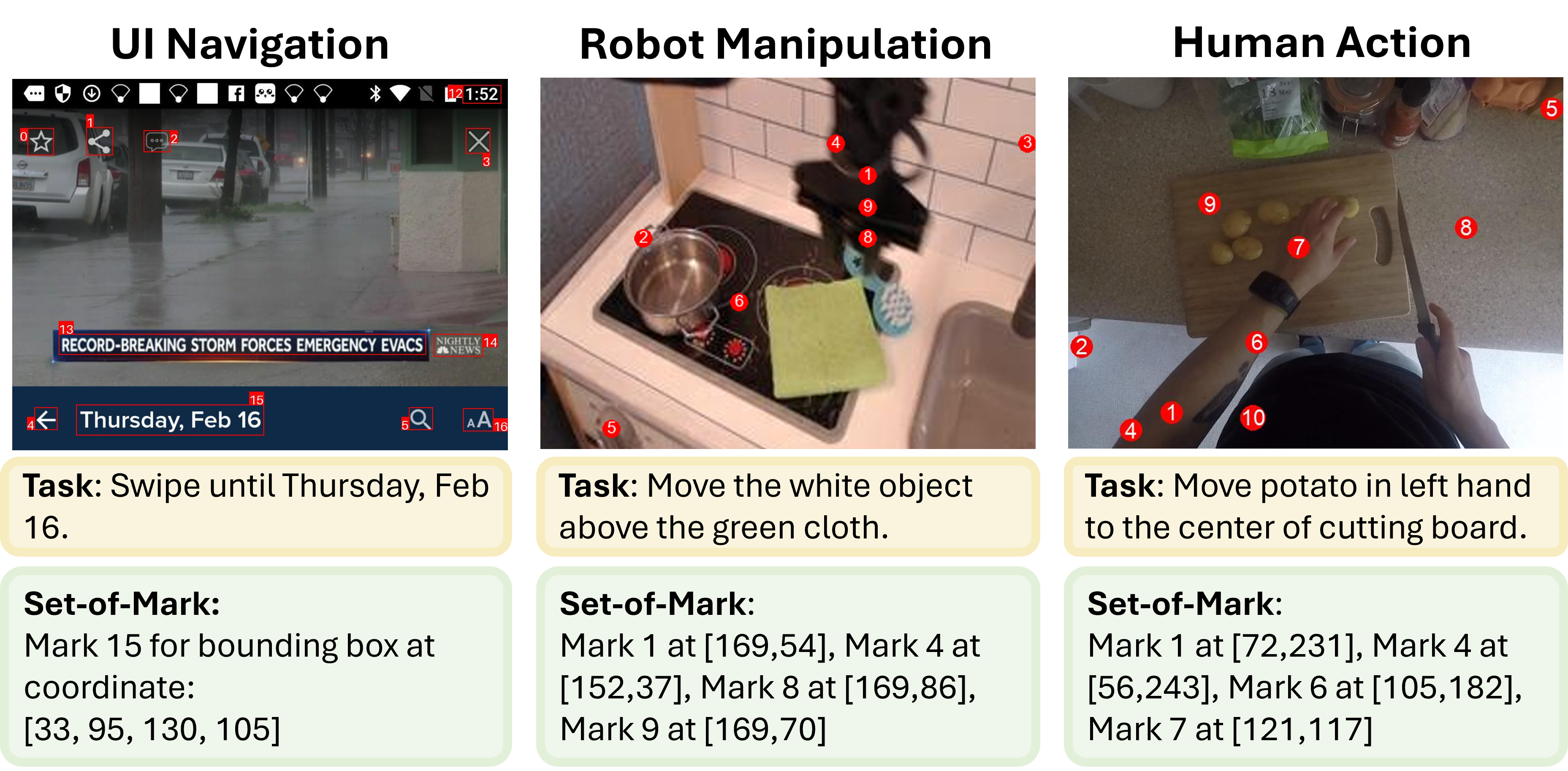}
    \caption{Set-of-Mark supervisions for action grounding on UI screenshot~(left), robot manipulation~(middle) and human video~(right). All coordinates are normalized by image size~(height, width) and then quantized into 256 bins. Images better viewed by zooming in.}
    \label{fig:som_illustration}
\end{figure}
Given an image observation $I_t\in \mathcal{R}^{H \times W \times 3}$ at timestep $t$, a task $\texttt{task}$ and context $\texttt{ctx}$, we first extract a set of $K$ candidate regions or points that are actionable $\mathcal{P} = \{p_1,...,p_K\}$, where $p_k$ could be a four-dimensional box coordinate or two-dimensional point coordinates. Subsequently, we overlay the marks and boxes (if any) to the corresponding location of the image with numerical labels, \ie, $\mathcal{M} = \{1:p_1,2:p_2,...,K:p_K\}$ giving us a new marked image $I_t^{M}$.

Given the prompted image $I_t^{M}$ in an atomic action step, the model needs to select the candidate marks along with the original coordinates, significantly easing the action grounding for the agentic model. In this way, Eq.~\eqref{equ1:original target} can be reformulated as:
\begin{equation}
    {o}_t^{mark} = \texttt{action}_t:\texttt{mark}_t = \bm{\pi}(\mathcal{I}_t^{M}, \texttt{task}, \texttt{ctx})
    \label{equ2:reformulate with marker}
\end{equation}
where ${o}_t^{mark}$ is a subset of marks $\mathcal{M}$.

In Fig.~\ref{fig:som_illustration}, we show a few instances to demonstrate the SoM-based action grounding in Fig~\ref{fig:teaser}. 
To obtain candidate regions to mark, we can leverage different proposal networks such as image segmentation models~\cite{zou2023segment,kirillov2023segment}, object detection models~\cite{liu2023grounding,li2021grounded}, or domain-specific models~\cite{lu2024omniparser}. Readers refer to Supp. for more details.

\subsubsection{Trace-of-Mark for Action Planning}
\label{sec:tom}

\begin{figure*}[t]
    \centering
    \includegraphics[width=1.0\linewidth]{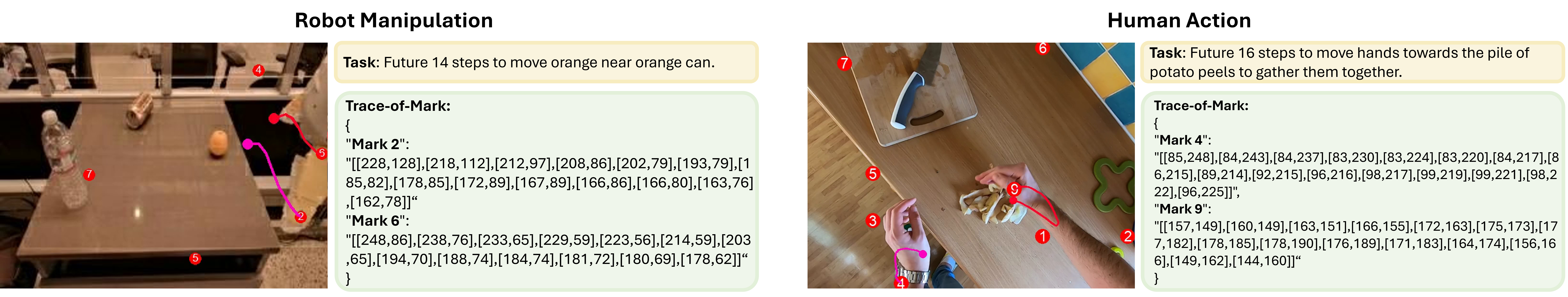}
    \vspace{-0.7cm}
    \caption{Trace-of-Mark supervisions for robot manipulation~(left) and human action~(right). Same coordinate normalization and quantization is used as SoM. Images show the future traces to predict.}
    \label{fig:tom_illustration}
\end{figure*}
Video data contains a lot of information about human actions and activities, which can essentially be leveraged to boost the capability of agentic models. However, due to the absence of action labels, previous methods rarely explore this direction, apart from a few works focused on world model learning~\cite{mendonca2023structured,liu2024world}.
We extend the strategy of ``overlaying marks'' from static images to dynamic videos by proposing Trace-of-Mark~(ToM) to allow the agentic model to effectively learn to plan and act from videos. 

Given the sequence of visual observations from a video $\mathcal{I} = \{I_1,...,I_t\}$, we extend along the time axis to the future $l$ frames, $\mathcal{I}_{future} = \{I_{t+1}, ..., I_{t+l}\}$. Given the $K$ marks at $t$-th frame $I_t$, we extract the corresponding positions of the overlay marks in the next $l$ frames, denoted traces $\mathcal{T} = \{\mathcal{M}_{t+1}, ..., \mathcal{M}_{t+l}\}$. Following the prediction of action type and valid marks as in Eq.~\eqref{equ2:reformulate with marker}, we further ask the model to predict the future trajectories for the valid marks:
\begin{equation}
\begin{aligned}
  {o}_t^{mark} & = \texttt{action}_t:\texttt{mark}_t:\texttt{trace}_{t+1:t+l} \\
& = \pi(\{\mathcal{I}_1,...,\mathcal{I}_{t-1},\mathcal{I}_t^{M}\}, \texttt{task}, \texttt{ctx})    
\end{aligned}
\label{EQ:ToM}
\end{equation}
where $\texttt{trace}_{t+1:t+l}$ is a subset of the trace sequences for valid marks in $\texttt{mark}_t$ in $\mathcal{T}$. Our proposed ToM predicting is a simple yet effective way of leveraging video data and brings two unique modeling benefits: $(i)$ It forces the model to understand the temporal dynamics in the video observations and to ``look ahead of time'' before taking the next actions; $(ii)$ Unlike predicting next frames as used in~\cite{liu2024worldmodelmillionlengthvideo}, predicting traces uses much fewer tokens to capture much longer temporal horizon and action-related object dynamics, while disregarding ambient contents.

To extract ToM, we employ point tracking models CoTracker~\cite{karaev2023cotracker}, though any performant model can be used. In particular, given a sequence of frames $\{I_t,I_{t+1},...,I_{t+l}\} \in \mathcal{R}^{(l+1) \times H\times W \times 3}$, we apply a dense tracking for $s^2$ grid points to extract $s^2$ traces of length $(l+1)$. Given these $s^2$ traces, we drop those traces whose average motion magnitudes between two adjacent timesteps are smaller than a certain value $\epsilon$ (Please see more details in the supplementary material). The remaining ones are regarded as foreground motions driven by a given task.

\subsection{Modeling}

To retain the multimodal understanding capability required for \magma, we adopt the common practice used in current VLMs~(\eg, LLaVA~\cite{liu2023llava} and Phi-3-Vision~\cite{abdin2024phi3}). Given the visual observations $\mathcal{I}$, we use a vision encoder $\mathcal{V}$ to encode each frame into a number of tokens and then concatenate all tokens into a sequence and feed them to a decoder-only LLM along with the language tokens that encode task descriptions. Due to the task diversity, a vision encoder that can seamlessly encode images and videos of various resolutions is needed. In light of this, we propose to use convolutional networks ConvNeXt~\cite{liu2022convnet} as the vision backbone, considering that it supports arbitrary image resolutions by default. To handle the high-resolution images (\eg, up to 2000 for UI screenshots), we simply perform global encoding without the bells and whistles used in previous work and find that it can encode the global context as well as combining global and local crops~\cite{liu2024llavanext,abdin2024phi3}. 
To that end, we formulate the agentic modeling as an autoregressive decoding procedure:
\begin{equation}
    o^{l,*}_{t+1} \sim p(o_{t+1}^l | \{o_1^l,...,o_t^l\};  \mathcal{V}(\mathcal{I}), \texttt{task}, \texttt{ctx}).
\end{equation}

\section{Multimodal Agentic Pretraining}

\subsection{Datasets}
\label{sec:pretraining_data}
To develop a foundation model with both verbal and spatial intelligence that is capable of handling diverse agentic tasks, we curated a comprehensive pretraining dataset from a wide range of images, videos, and robotics domains.

\begin{table}[t]
    \centering
    \begin{tabular}{l|cc}
        Data Type &  Set-of-Mark & Trace-of-Mark\\
        \toprule
        UI Screenshots & \checkmark & \xmark \\
        Robotics Images & \checkmark & \checkmark \\
        Instructional Videos & \checkmark & \checkmark \\
    \end{tabular}
    \vspace{-0.2cm}
    \caption{SoM and ToM applied to various data types. ToM is not applied to UI data as they are a sequence of discrete screenshots.}
    \label{tab:som_tom_generation}
\end{table}

\begin{itemize}
\item \textbf{Robotics manipulation data}. For robotics task, we follow OpenVLA~\cite{kim2024openvla} and use the robotics dataset of Open-X-Embodiment~\cite{embodimentcollaboration2024openxembodimentroboticlearning}.

\item \textbf{UI navigation Data}. We exploit two pretraining datasets, SeeClick~\cite{seeclick} and Vision2UI~\cite{gui2024vision2uirealworlddatasetlayout}. 
\item \textbf{Instructional videos}. We compile Epic-Kitchen~\cite{Damen2018EPICKITCHENS, Damen2022RESCALING}, Ego4d~\cite{grauman2022ego4dworld3000hours}, Somethingv2~\cite{goyal2017something} and other related considering the coarse but rich goal-driven human actions.
\item \textbf{Multimodal understanding}. Lastly, we include ShareGPT4V~\cite{chen2023sharegpt4vimprovinglargemultimodal}, instruction tuning data in LLaVA-1.5~\cite{liu2024llavanext}, and a few other OCR-related datasets~\cite{masry-etal-2022-chartqa,mathew2021infographicvqa} to attain image understanding capability.
\end{itemize}
We noticed that many more related datasets could be used for our model pretraining, such as large-scale instruction tuning data~\cite{tong2024cambrian,li2024llavaonevision}, more diverse video data~\cite{chen2024panda70mcaptioning70mvideos}. In this study, we focus on the demonstration of our pretraining methodology and leave the further scaling up for future. In the next, we elaborate on how we extract the surrogate action supervisions through Set-of-Mark~(SoM) and Trace-of-Mark~(ToM).

\subsection{SoM and ToM Generation}

As shown in Table~\ref{tab:som_tom_generation}, we apply SoM and ToM for different data types, where SoM is applied to all to learn a uinified action grounding. ToM is not fit for the UI data as it consists of sequences of discrete screenshots.

\subsubsection{SoM for UI Navigation}

For UI screenshots in our pretraining data, we mainly rely on the original annotations extracted based on DoM Tree. In addition to the bounding boxes extracted from HTML code~\cite{seeclick,gui2024vision2uirealworlddatasetlayout}, we further annotate the mobile screenshots in SeeClick data with bounding boxes derived from Android view hierarchies~\cite{rico_semantics}. 
Given the extracted candidate bounding boxes for an image, we apply Alg.~\ref{alg:som_prompting} to assign a textual label (line 3) and draw the boxes around the objects. To minimize overlapping box placements, we determine the optimal position for a label using previously drawn boxes (line 5) before computing the textbox size and assigning its coordinates (line 7). During the evaluation, we follow the common practice by applying OmniParser~\cite{lu2024omniparser} for the zero-shot evaluation on ScreenSpot~\cite{seeclick}, and using the candidate boxes provided by~\cite{mind2web} for downstream training and evaluation on Mind2Web.
\begin{algorithm}[t]
\caption{SoM generation for UI images}
\label{alg:som_generation}
\begin{algorithmic}[1]
    \Require image $I$, bounding boxes $B$, image height and width $(i_h, i_w)$
    \State $B^* \gets []$ 
    \For{$(idx, b) \in \texttt{enumerate}(B)$}  
        \State $text \gets \texttt{str}(idx+1)$
        \State $I \gets \texttt{DrawRectangle}(I, b)$
        \State $(c_y, c_x) \gets \texttt{FindOptimalCorner}(b, B^*, (i_h, i_w))$
        \textcolor{commentcolor}{\Comment{Find corner that is far away from all boxes in $B^*$}}
        \State $(m_h, m_w) \gets \texttt{GetMarkSize}(text, H, W)$        
        \State $text\_box \gets (c_y, c_x, c_y \pm m_h, c_x \pm m_w)$
        \State $I \gets \texttt{DrawRectangle}(I, text\_box)$
        \State $I \gets \texttt{DrawText}(I, (c_x, c_y), text, color=white)$
        
        \State $B^* \gets B^* + [b]$ \textcolor{commentcolor}{\Comment{Add current drawn box to $B^*$}}
    \EndFor
    
    \State \textbf{Return} $I$
\end{algorithmic}
\label{alg:som_prompting}
\end{algorithm}

\subsubsection{SoM and ToM for Videos and Robotic Data}
\label{sec:main_tom}

We use marks and traces as surrogate action supervisions to pretrain our \magma model for action grounding and planning. To extract reliable traces, we use the state-of-the-art point tracking model CoTracker~\cite{karaev2023cotracker} to track the keypoints in each video segment. Unlike object detection and tracking systems used in previous works~\cite{ravi2024sam2segmentimages,niu2024llarva,li2025hamsterhierarchicalactionmodels}, point tracking provides the finest grained moving trajectories for both end effectors~(robot arms or human hands) and objects, and more importantly can be feasibly applied to any videos as it does not require object recognition.

\begin{algorithm}[t]
\caption{SoM and ToM generation for instructional videos and robotic data}
\label{alg:som_tom_generation}
\begin{algorithmic}[1]
    \Require image sequence $\mathcal{I} = \{I_t,...I_{l}\}$; grid size $s$; global motion threshold $\eta$; foreground threshold $\epsilon$
    \State $\mathcal{M} = \{M_t,...,M_{l}\} \gets \texttt{CoTracker}(\mathcal{I}, s)$
    \If{$\texttt{HasGlobalMotion}(\mathcal{M}, \eta)$}
      \State $\mathcal{M} \gets \mathcal{H}(\mathcal{M})$ \textcolor{commentcolor}{\Comment{Apply homography transformation}}
    \EndIf
    \State $\mathcal{M}^f, \mathcal{M}^b = \texttt{ClassifyTraces}(\mathcal{M}, \epsilon)$\textcolor{commentcolor}{\Comment{Classify traces into foreground and background ones}}
    \State $k \gets \texttt{Random}(1, \min(5, |\mathcal{M}^f|))$ 
    \State $\mathcal{M}^f,\mathcal{M}^b = \texttt{KMeans}(\mathcal{M}^f, k), \texttt{KMeans}(\mathcal{M}^b, 2k)$    
    \textcolor{commentcolor}{\Comment{Cluster foreground and background traces separately}}
    \State $I_t \gets SoM(I_t, \{M^f_t, M^b_t\})$ \textcolor{commentcolor}{\Comment{Apply SoM on 1st frame}}
    \State \textbf{Return} $\mathcal{I}, \mathcal{M}_f^*$
\end{algorithmic}
\label{alg:tom_prompting}
\end{algorithm}
\begin{figure}[t]
    \centering
    \includegraphics[width=1.0\linewidth]{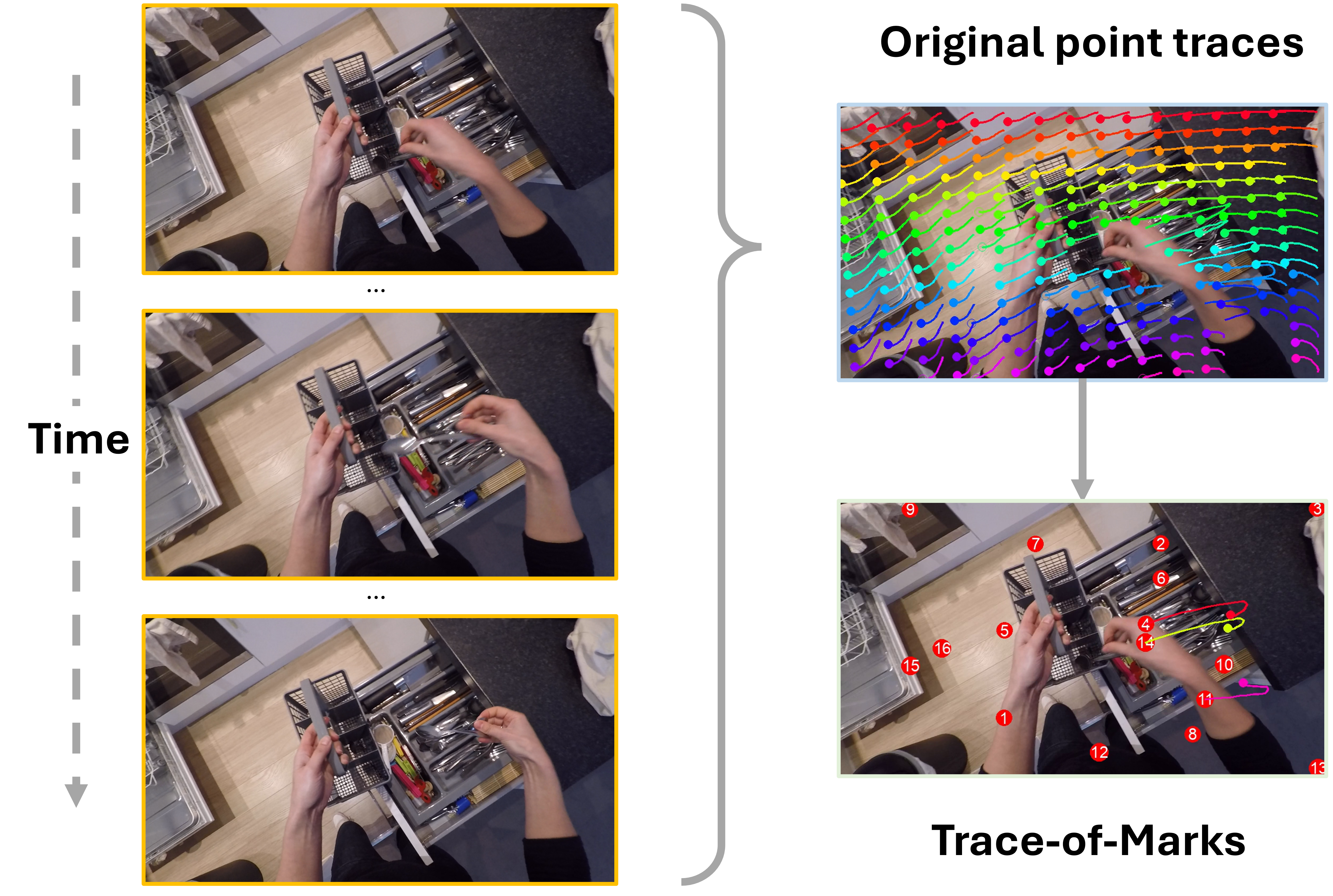}
    \vspace{-10pt}
    \caption{An illustration of Alg.~\ref{alg:tom_prompting} to handle videos with camera motions for SoM/ToM generation.}
    \label{fig:homo}
\end{figure}
\begin{figure*}[!t]
    \centering
    \begin{minipage}{0.29\textwidth}
        \centering
        \includegraphics[width=\textwidth]{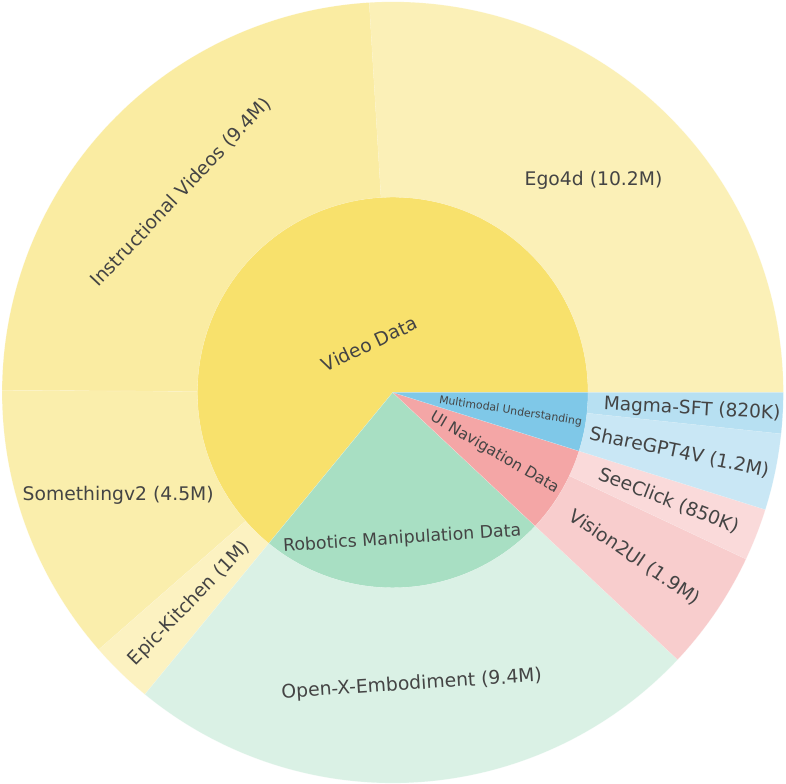} 
    \end{minipage}
    \hfill
    \begin{minipage}{0.7\textwidth}
        \centering
        \includegraphics[width=\textwidth]{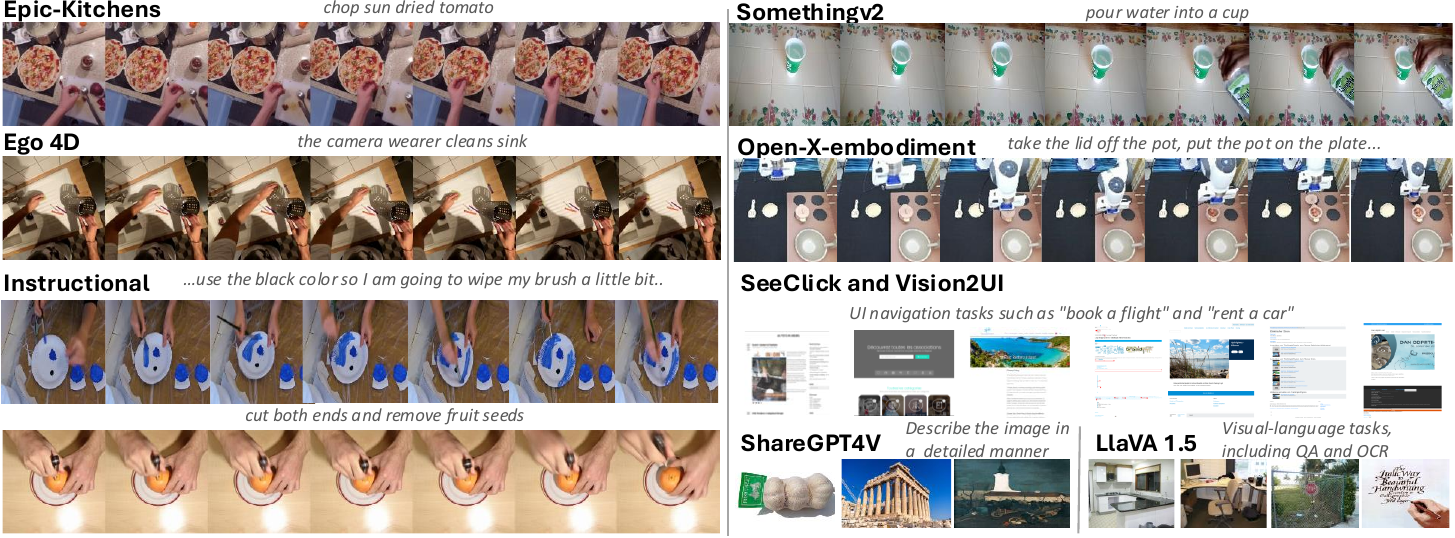}
    \end{minipage}
    \caption{\textbf{Overview of Pretraining Data Sources.} A diverse collection of datasets including instructional videos (\textcolor[RGB]{242,150,75}{orange}), robotics manipulation (\textcolor[RGB]{102,188,156}{green}), UI navigation (\textcolor[RGB]{235,150,170}{pink}), and multimodal understanding (\textcolor[RGB]{70,130,180}{blue}). Note that we count the size of each dataset by the number of image samples. For video and robotics data, we extract the images from the short clips and trajectories, respectively.}
    \label{fig:dataset_vis}
\end{figure*}

\noindent\textbf{Reliability of CoTracker}. To determine the generalizability of such traces, we examine the reliability of CoTracker before running the algorithm on all our pretraining data. We note that CoTracker was already well validated on multiple video datasets such as TAP-Vid~\cite{doersch2022tap} and PointOdyssey~\cite{zheng2023pointodyssey} in the original paper. In this work, we proposed comprehensive strategies to handle scene transition and camera motions in videos~(Alg.~\ref{alg:tom_prompting}), which effectively scale to datasets like Ego4D and other instructional videos~(Fig~\ref{fig:data_dist}). To further validate the reliability of ToM, we quantitatively evaluated the traces on a subset of YouCook2-BB~\cite{ZhLoCoBMVC18} with box annotations by humans. We extract the traces from each annotated box and count the number of future traces still falling into the box 1 second forward. On 1320 clips, we got a precision of \textbf{0.89}, indicating that the traces reliably capture temporal motions. 

\noindent\textbf{Segment and CLIP-score filtering} As the point tracking system works in a short time window, we begin by using the annotations provided, curated or otherwise, to split each video into segments, and then run PySceneDetect~\cite{PySceneDetect} to further break each segment into short video clips with consistent shots. However, the detected video clips may not always be relevant to their associated text annotations. Thus, we use the pretrained CLIP~\cite{radford2021learning} visual and text encoders to compute the cosine similarity score between each clip and text pair, and filter out clips with $<0.25$ scores.

Once we have the fine-grained video clips in hand, we apply Alg.~\ref{alg:tom_prompting} to generate SoM and ToM. Given a video clip with $l$ frames $\{I_1,I_{2},...,I_{l}\} \in \mathcal{R}^{(l) \times H\times W \times 3}$, we start from the time step $t$  and put a grid of equally spaced $s^2$ points on $I_t$. Then, we use CoTracker to extract $s^2$ future traces of length $(l-t)$ each. The output also contains predicted occlusion labels for each trace, which indicate if any points on the trace are obstructed at some time steps. 

\noindent\textbf{Removal of global motions}. Many instructional videos, particularly the ego-centric ones~\cite{grauman2022ego4dworld3000hours}, contain significant camera movements. Consequently, the extracted traces may reflect external movements instead of relevant actions to accomplish a given task. We mitigate this issue by performing the homography transformation~\cite{dubrofsky2009homography}. Specifically, we compute the $3\times 3$ transformation matrix $h_i$ with the future mark positions and current ones:
\begin{equation}
    h_i = \mathcal{H}({M}_t, {M}_{t+i}) \in \mathcal{R}^{3\times 3}
\end{equation}
Given $h_i$, we apply the homography transformation to $M_{t+i}$ to obtain $M^*_{t+i}$ which shares the same coordinate system as $M_{t}$. Valid traces of marks to predict in Eq.~\eqref{EQ:ToM} are then extracted from $\{M_{t}, M^*_{t+1}, M^*_{t+l}\}$. It turns out that the proposed method is effective to remove global camera motions for both ego-centric videos and exo-centric ones, as ilustrated in Fig.~\ref{fig:homo}.

After extracting the traces and applying the homography transformation if needed (lines 2-4), we classify them into two categories, foreground and background traces based on the average motion magnitude between two adjacent time steps, where traces with average motion magnitude of at least $\epsilon$ (line 5) are counted as foreground. Finally, we select the number of clusters (line 6) and perform a K-Means clustering for the foreground and background traces separately~(line 7) before randomly selecting one or more points from each cluster as the final traces. In practice, we set $s$, $\eta$ and $\epsilon$ to be 15, 2 and 2, respectively.

\subsection{Pretraining}

\begin{figure*}
    \centering
    \includegraphics[width=1.0\linewidth]{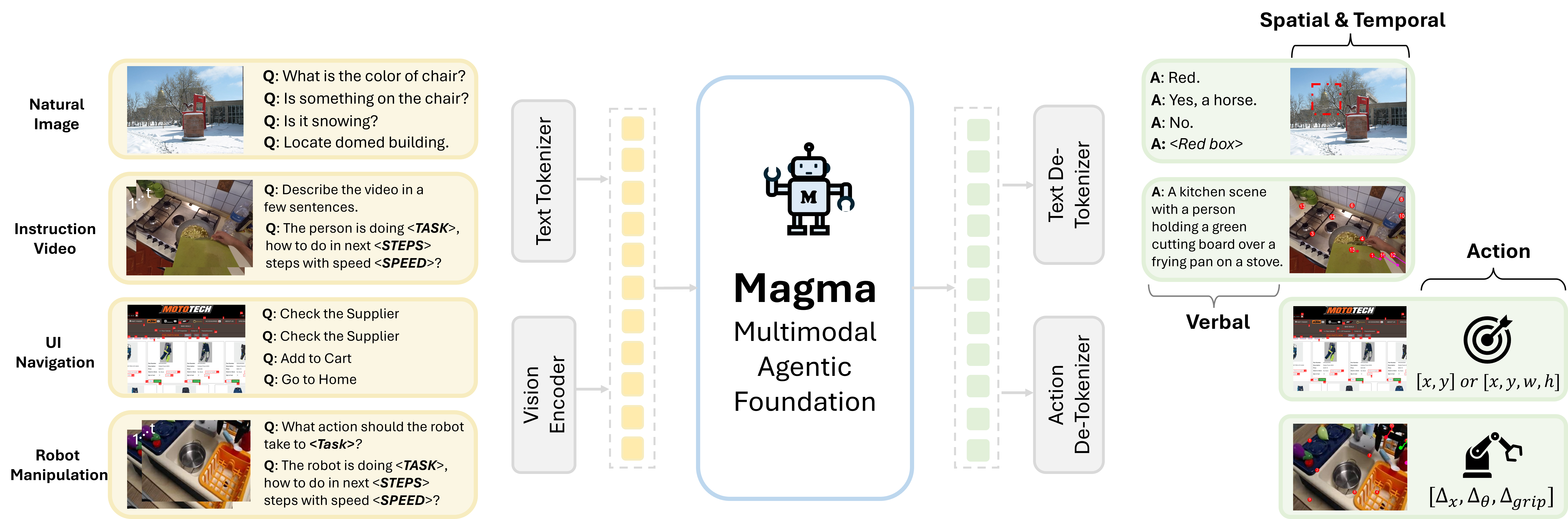}
    \vspace{-0.5cm}
    \caption{Magma pretraining pipeline. For all training data, texts are tokenized into tokens, while images and videos from different domains are encoded by a shared vision encoder. The resulted discrete and continuous tokens are then fed into a LLM to generate the outputs in verbal, spatial and action types. Our proposed method reconcile the multimodal understanding and action prediction tasks.}
    \label{fig:magma_pt}
\end{figure*}

The above data and annotation curation results in a comprehensive pretraining suite which covers $(i)$ different digital and physical environments; $(ii)$ both verbal and spatial annotations and $(iii)$ various multimodal understanding and agentic tasks. As seen in Fig.~\ref{fig:dataset_vis}~(left), we include close to 2.7M UI navigation screenshots from SeeClick~\cite{seeclick} and Vision2UI~\cite{gui2024vision2uirealworlddatasetlayout}. We follow OpenVLA~\cite{kim2024openvla} to incorporate 970K trajectories in Open-X-Embodiment~\cite{open_x_embodiment_rt_x_2023}, which consists of 9.4M image-
language-action triplets. Another majority of the pretraining data are videos which comprise over 25M samples sourced from around 4M shot-consistent video clips. 
Finally, we include 1.2M image and text pairs from ShareGPT4V~\cite{chen2023sharegpt4v}, and LLaVa-1.5~\cite{liu2024improvedbaselinesvisualinstruction} and a few other OCR-related datasets~\cite{masry-etal-2022-chartqa,mathew2021infographicvqa}, which we denote by \textit{ Magma-SFT}~(820K). 

By default, we use LLaMA-3-8B~\cite{llama3} as the language backbone and ConvNext-XXlarge~\cite{liu2022convnet} as the vision backbone. We show the pretraining architecture in Fig.~\ref{fig:magma_pt}. Our proposed SoM and ToM play as the bridge to connect verbal and action supervisions for all four types of data, and significantly enhance model's spatial intelligence as we observe during our experiments.

For comparisons, we run a few variants for the ablation studies in our experiments:
\begin{itemize}
    \item {\magma-8B~(SFT)} is the model trained with \textit{Magma-SFT}~(820K) for the instruction tuning following a conventional recipe used on LMM training. 
    \item {\magma-8B~(UI)} and {\magma-8B~(OXE)} are the models pretrained on UI screenshots and OXE robotics data, respectively.
    \item {\magma-8B~(ACT)} is pretrained jointly on UI screenshots and robotics data.
    \item {\magma-8B~(Full)} is the full model trained with the whole dataset with SoM and ToM annotations.
\end{itemize}

Unless noted otherwise, all pretrainng includes the \textit{Magma-SFT}~(820K). We pretrain our model using our curated data for maximally three epochs with a constant learning rate of 1e-5, and evaluate the pretrained model on different tasks under the zero-shot setting as well as finetune its weights on the downstream tasks. The entire model including the parameters of the language model and the vision encoder is tuned. See Appendix for more detailed settings.
\section{Experiment}
\begin{table*}[ht]
    \centering
    \resizebox{1.0\linewidth}{!}{
    \begin{tabular}{lc|ccccccccccccccc}
    &   & \multicolumn{3}{c}{\rotatebox{0}{\textbf{Multimodal Understanding}}} & \multicolumn{5}{c}{\rotatebox{0}{\textbf{UI Action Grounding and Navigation}}} & \multicolumn{2}{c}{\textbf{Robot Manipulation}} \\
    \toprule
    Model  & Size&  VQAv2 & TextVQA & POPE & \textit{SS}-Mobile & \textit{SS}-Desktop & SS-Web & VWB-Ele-G & VWB-Act-G &  SE-Google Robot & SE-Bridge \\
    \midrule
    {GPT-4V}~\cite{gpt4v} & {n/a} & 77.2 & \textbf{78.0} & {n/a} & {22.6/24.5} & {20.2/11.8} & {9.2/8.8} & {\underline{67.5}} &{\textbf{75.7}} & - & - \\  
    {GPT-4V-OmniParser}~\cite{lu2024omniparser} & {n/a} & n/a & n/a & {n/a} & {\textbf{92.7}/49.4} & {64.9/26.3} & {\textbf{77.3}/39.7} & - & - & - & -\\
          \hline
          \hline
    LLaVA-1.5~\cite{liu2023llava} & 7.4B &78.5 & 58.2 & 85.9 & - & - & - & 12.1 & 13.6 & - & -\\
    LLaVA-Next~\cite{liu2024llavanext} & 7.4B & \textbf{81.8} & 64.9 & \underline{86.5} & - & - & - & 15.0 & 8.7 & - & - \\    
    Qwen-VL~\cite{Qwen-VL} & 9.6B & 78.8 & 63.8 & n/a & 7.5/4.8 & 5.7/5.0 & 3.5/2.4 & 14.0 & 10.7 & - & -\\
    Qwen-VL-Chat~\cite{Qwen-VL} & 9.6B & 78.2 & 61.5 & n/a & - & - & - & - & - & - & -\\
    
    \midrule
    Fuyu~\cite{fuyu_8b} & 8B  & 74.2 & n/a & n/a & 41.0/1.3 & 33.0/3.6 & 33.9/4.4 & 19.4 & 15.5 & - & -\\    SeeClick~\cite{seeclick} & 9.6B & - & - & - & \underline{78.0}/\underline{52.0} & 72.2/\underline{30.0} & 55.7/32.5 & 9.9 & 1.9 & - & -\\
Octo~\cite{team2024octo} & 93M  & -&-&-&- & - & - &   - & -     &   6.0 & \underline{15.9}\\
    RT-1-X~\cite{open_x_embodiment_rt_x_2023} & 35M  & - & - & - & - & - &  -   & -     & -   &   \underline{34.2} & 1.1 \\
    OpenVLA~\cite{kim2024openvla} & 8B  & - & - & - & - & - &  -   & -     & -   &   31.7 & 14.5 \\
    \midrule
    {\magma-8B~(Ours)} & 8.6B & \underline{80.0} & \underline{66.5} & \textbf{87.4} &60.4/\textbf{58.5} & \textbf{75.3/52.9} & {69.1}/\textbf{52.0} & \textbf{96.3} & \underline{71.8}  & \textbf{52.3} & \textbf{35.4} \\    
    \end{tabular}
    }
    \vspace{-3pt}
    \caption{\textbf{Zero-shot evaluation on agentic intelligence}. We report the results for pretrained \magma \textit{without} any domain-specific finetuning. \magma is the only model that can conduct the full task spectrum. ``SS'' denotes the ScreenSpot benchmark proposed in SeeClick~\cite{seeclick}; ``VWB'' denotes VisualWebBench~\cite{liu2024visualwebbench}; ``SE'' denotes the SimplerEnv simulator~\cite{li24simpler}. `n/a' means not available and `-' means not supported. For all related evaluations, we use OmniParser to provide the detection results only, without local semantics.}
\label{tab:agentic_evaluation}
\end{table*}

\begin{figure*}[!t]
    \centering
    \includegraphics[width=\textwidth]{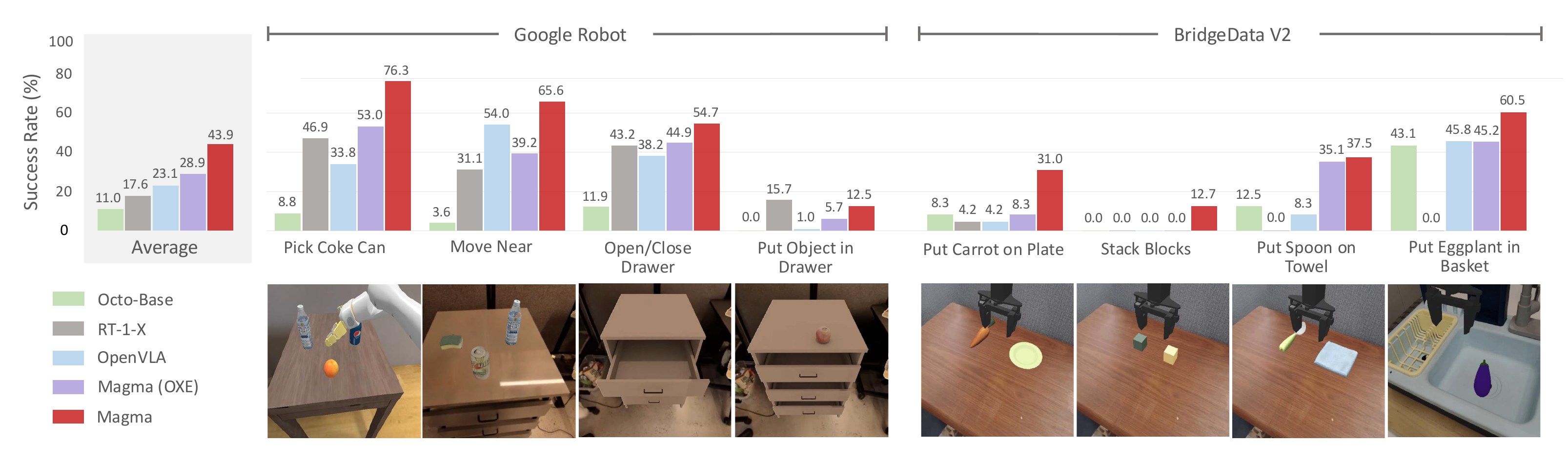}
    \vspace{-0.6cm}
    \caption{\textbf{SimplerEnv performance comparison on Google Robots and Bridge.} \texttt{Magma}(OXE) represents our model trained solely on Open-X-Embodiment (OXE)~\cite{embodimentcollaboration2024openxembodimentroboticlearning}, while \texttt{Magma} is our pretrained model. Results for each task are averaged across visual matching and variant aggregation scenarios.}
    \label{fig:simpler_env}
\vspace{-10pt}
\end{figure*}

\subsection{Evaluating Agentic Capability}

We examine the effectiveness of \magma as the foundation model for multmodal agents on UI Navigation tasks in the digital world, the robotic manipulation in the physical world, as well as the generic multimodal understanding.

\subsubsection{Zero-Shot Evaluation}

To evaluate the zero-shot transferability of \magma,  we employ ScreenSpot~\cite{seeclick} and VisualWebBench~\cite{liu2024visualwebbench} for evaluating UI action grounding and navigation, and SimplerEnv~\cite{li24simpler} for robotic manipulation. In addition to these evaluations, we also validate our model on generic~\cite{goyal2017vqav2} and text-rich~\cite{singh2019textvqa} VQA tasks as well as hallucination benchmark POPE~\cite{li2023pope}. As shown in Table~\ref{tab:agentic_evaluation}, \magma~consistently outperforms all other general-domain LMMs~(\eg, LLaVA, Qwen-VL) and domain-specific agentic models such as SeeClick~\cite{seeclick} for UI navigation and OpenVLA~\cite{kim2024openvla} for robotic manipulation. Notably, the zero-shot performance of \magma on UI is much better than the state-of-the-art vision-based method that uses GPT-4V and Omniparser~\cite{lu2024omniparserpurevisionbased}. 
We report the results on two commonly used simulator embodiments in SimplerEnv~\cite{li24simpler}, Bridge and Google Robot including 8 tasks with 172 visual matching and variant aggregation scenarios. 
Since OpenVLA uses real robot trajectories for pre-training, the model is susceptible to the domain gap for real-to-sim adaptation. In contrast, our \magma model, trained for multimodal understanding and action prediction on a wide range of heterogeneous datasets, is significantly more resilient to the gap and achieves significantly better success rates. 


Fig.~\ref{fig:simpler_env} shows detailed comparisons between our pretrained \magma model and other representative models. Remarkably, \magma surpasses the second-place OpenVLA by \textbf{19.6}\%, nearly doubling the average success rate. On those challenging tasks such as ``Put Object in Drawer'' and ``Put Carrot on Plate'', \magma achieves a remarkable success rate while most baselines fail entirely. Notably, \magma tuned on our pretrained model showcases substantially better results than the version trained solely on robotic datasets, highlighting the value of spatial intelligence learned from diverse datasets for physical robotic manipulation tasks.

\begin{table}[t]
\vspace{-4pt}
    \centering
    \resizebox{1.0\linewidth}{!}{
    \begin{tabular}{lc|ccccc}
    \toprule
    Model & SoM+ToM & SS-Overal & VWB-Ele-G & VWB-Act-G& SE-Bridge & SE-Google\\
    \midrule
    \magma-8B~(UI) & \xmark & 57.7 & 68.5 & 58.3 & - & -\\
    \magma-8B~(OXE) & \xmark & - & - & - & 22.2 & 35.7 \\
    \magma-8B~(ACT) & \xmark & {56.2} & 89.1 & 21.4 & 17.5 & 31.5\\
    \magma-8B~(Full) & \xmark & 57.4 & {90.1} & {25.2} & {17.7} & {37.5} \\
    \magma-8B~(Full) & \cmark & \textbf{61.4} & \textbf{96.3} & \textbf{71.8} & \textbf{35.4} & \textbf{52.3} \\
    \bottomrule
    \end{tabular}}
    \vspace{-10pt}
    \caption{Ablation study on the effect of data mixtures and pretraining techniques. w/o SoM+Tom means using original action supervisions~(2D coordinates for UI and 7DoF for robots.)}
\label{tab:ablations}
    \vspace{-8pt}
\end{table}

\noindent\textbf{Ablation Studies.} We ablate our model pretraining techniques and data mixtures. The results are shown in Table~\ref{tab:ablations}. First, we observe from the top three rows that simply combining UI and robotics data does not bring gains, but instead hurts the performance for both tasks. This is expected because the two agentic tasks have significantly different image domains as well as action spaces~(2D coordinates \textit{v.s.} 7-DoF). Adding video data to the pretraining slightly improves the performance across board but still can not fill the gap in between, as the additional video narrations can only enhance the verbal intelligence. However, once we apply SoM and ToM to all the pretraining data to put them into the unified interface, our model can learn effectively from the heterogeneous data for both verbal and spatial intelligence. This study highlights the effectiveness of our proposed method and indicates equally importance of verbal and spatial understanding for agentic tasks.


\subsubsection{Efficient Finetuning}

\begin{table*}[]
    \centering
    \footnotesize
    \resizebox{\textwidth}{!}{%
    \begin{tabular}{l|lcccccccccccccccc}
        \multirow{2}{*}{Method} &  \multirow{2}{*}{Backbone} & \multicolumn{2}{c}{Input Source} & \multicolumn{3}{c}{Cross-Website}  & \multicolumn{3}{c}{Cross-Task} & \multicolumn{3}{c}{Cross-Domain} \\
         & & DoM Tree & Image &  Ele. Acc & Op. F1 & Step SR & Ele. Acc & Op. F1 & Step SR & Ele. Acc & Op. F1 & Step SR  \\        
        \toprule
         \multirow{1}{*}{GPT-4-MindAct~\cite{mind2web}} & GPT-4~\cite{gpt4} & \cmark & & 35.8 & 51.1 & 30.1 & 41.6 & 60.6 & 36.2 & 37.1 & 46.5 & 26.4 \\
         \multirow{1}{*}{GPT-4V-OmniParser~\cite{lu2024omniparser}} & GPT-4V~\cite{gpt4v} & \cmark & \cmark & 41.0 & \textbf{84.8} & 36.5 & 42.4 & \textbf{87.6} & 39.4 & 45.5 & \textbf{85.7} & 42.0 \\
        \midrule 
        \multirow{3}{*}{SeeAct~\cite{zheng2024gpt4vision}} & GPT-4V~\cite{gpt4v} &  & \cmark & & & 13.9 & - & - & 20.3 & - & - & 23.7 \\
        & Gemini-Pro~\cite{geminiteam2024gemini} & \cmark & \cmark & 21.5 & 67.7 & 19.6 & 21.5 & 67.7 & 19.6 & 20.7 & 64.3 & 18.0\\
        & GPT-4V~\cite{gpt4v} & \cmark & \cmark & 38.0 & 67.8 & 32.4 & 46.4 & 73.4 & 40.2 & 42.4 &69.3 &36.8\\
          \hline
          \hline
          Fuyu-8B$^{\ddag}$ & Fuyu-8B~\cite{fuyu_8b} & & \cmark  & 4.8 & 81.3 & 4.0 & 8.3 & 83.9 & 6.6 & 3.6 & 83.0 & 3.0\\
          Fuyu-8B-GUI~\cite{chen2024guicourse} & Fuyu-8B~\cite{fuyu_8b} & & \cmark  & 13.9 & 80.7 & 12.2 & 19.1 & 86.1 & 15.6 & 14.2 & 83.1 & 11.7\\
          MiniCPM-V$^{\ddag}$ & MiniCPM-V~\cite{yao2024minicpm} & & \cmark  & 8.2 & 78.2 & 6.0 & 11.0 & 85.6 & 8.5 & 6.5 & 81.4 & 5.2\\
          MiniCPM-V-GUI~\cite{chen2024guicourse} & MiniCPM-V~\cite{yao2024minicpm} & & \cmark  & 20.3 & 81.7 & 17.3 & 23.8 & 86.8 & 20.8 & 17.9 & 74.5 & 17.6\\
          Qwen-VL$^{\natural}$ & Qwen-VL~\cite{Qwen-VL} & & \cmark  &13.2 & 83.5 &9.2 &15.9 &86.7 &13.3 & 14.1 & 84.3 & 12.0\\
          SeeClick~\cite{seeclick} & Qwen-VL~\cite{Qwen-VL} & & \cmark & 21.4 & 80.6 & 16.4 & 28.3 & 87.0 & 25.5 & 23.2 & 84.8 & 20.8 \\
          CogAgent$^{\dag}$~\cite{cogagent} & CogVLM~\cite{wang2023cogvlm} &  & \cmark & 27.3 & - & 23.4 & 30.2 & - & 26.9 & 33.1 & - & 28.5 \\
          Qwen2-UIX~\cite{multiUI} & Qwen2~\cite{Qwen2} & & \cmark & 39.2 & - & 31.0 & 43.4 & - & 38.2 & 40.4 & - & 34.9 \\     
        \midrule
          \magma-8B~(Ours) & LLaMA3~\cite{llama-3} & & \cmark & \textbf{57.2} & 76.9 & \textbf{45.4} & \textbf{54.8} & 79.7 & \textbf{43.4} & \textbf{55.7} & 80.6 & \textbf{47.3} \\
    \end{tabular}}
    \vspace{-5pt}
    \caption{\textbf{Efficient finetuning on Mind2Web for web UI navigation}. ``Ele. Acc'' denotes element selection accuracy. ``Op. F1'' denotes the token-wise F1 score between predicted ground-truth operation. ``Step SR'' denotes the step-wise success rate. $^{\ddag}$ Numbers reported in \citet{chen2024guicourse}. $^{\natural}$ Numbers reported in \citet{seeclick}. $^{\dag}$ Numbers reported in \citet{multiUI}.}
    \label{tab:mind2web}
\end{table*}

\begin{table*}[]
    \centering
    \footnotesize
    \resizebox{\textwidth}{!}{%
    \begin{tabular}{l|lcccccccccccccc}
         Method & Backbone & DoM Tree & Image &  General & Install & GoogleApps & Single & WebShopping & Overall \\        
        \hline
         \multirow{1}{*}{GPT-4V-SeeAct$^{\dag}$~\cite{zheng2024gpt4vision}} & GPT-4V~\cite{gpt4v} &  & \cmark & 34.1 & 39.4 & 40.0 & 46.2 & 38.2 & 39.6 \\
         \multirow{1}{*}{GPT-4V-ReAct$^{\dag}$~\cite{yao2022react}} & GPT-4V~\cite{gpt4v} &  & \cmark & 36.2 & 42.5 & 46.6 & 49.1 & 39.2 & 42.7 \\
         \multirow{1}{*}{GPT-4V-OmniParser~\cite{lu2024omniparser}} & GPT-4V~\cite{gpt4v} & \cmark & \cmark & 48.3 & 57.8 & 51.6 & 77.4 & 52.9 & 57.7 \\
         \hline
         \hline
          Fuyu-8B$^{\ddag}$ & Fuyu-8B~\cite{fuyu_8b} & & \cmark & - & 45.9 & 40.0 & 47.2 & 40.8 & - \\
          Fuyu-8B-GUI~\cite{chen2024guicourse} & Fuyu-8B~\cite{fuyu_8b}  & & \cmark & - & 50.9 & 41.6 & 45.7 & 43.8 & - \\
          MiniCPM-V$^{\ddag}$ & MiniCPM-V~\cite{yao2024minicpm} & & \cmark & - & 50.2 & 45.1 & 56.2 & 44.0 & - \\
          MiniCPM-V-GUI~\cite{chen2024guicourse} & MiniCPM-V~\cite{yao2024minicpm} & & \cmark & - & 62.3 & 46.5 & 67.3 & 57.5 & - \\
          Qwen-VL$^{\natural}$ & Qwen-VL~\cite{Qwen-VL} & & \cmark  & 49.5 & 59.9 & 46.9 & 64.7 & 50.7 & 54.3 \\
          SeeClick~\cite{seeclick} & Qwen-VL~\cite{Qwen-VL} & & \cmark & {54.0} & {66.4} & {54.9} & 63.5 & {57.6} & {59.3}\\
          
        \hline
          \magma-8B~(Ours) & LLaMA3~\cite{llama-3} & & \cmark & \textbf{61.5} & \textbf{73.2} & \textbf{62.7} & \textbf{77.5} & \textbf{61.7} & \textbf{67.3} \\
    \end{tabular}}
    \vspace{-5pt}
    \caption{\textbf{Efficient finetuning on AITW for mobile UI navigation}. We compared models either using DoM tree or image screenshot. We finetune our \magma jointly and then report the results on individual tasks. $^{\dag}$ Numbers reported in \citet{zhang2024dynamic}. $^{\ddag}$ Numbers reported in \citet{chen2024guicourse}. $^{\natural}$ Numbers reported in \citet{seeclick}.} 
    \label{tab:aitw}
\end{table*}

With moderate finetuning, the pretrained \magma model can be easily transferred to various downstream agentic tasks.

\noindent \textbf{{UI Navigation}}. Following the prior works~\cite{seeclick,cogagent}, we finetune \magma on Mind2Web and AITW, to examine the web and mobile UI navigation capabilities, respectively. For Mind2Web, we first apply the SoM prompting to the training samples according to the top candidates selected by~\cite{zheng2023seeact}, and then finetune \magma on the same samples as in SeeClick~\cite{seeclick}. Table~\ref{tab:mind2web} shows the results in three subtasks, and clearly indicates \magma's superiority to both general-domain and specific-domain LMMs. Similarly, on AITW \magma outperforms the state-of-the-art methods based on open-source or prosperity models. Considering that we use a similar size of LLM and a moderate amount of UI-related pretraining data, this decent performance is largely due to the proposed SoM and ToM modeling techniques, which significantly facilitate action grounding for UI navigation.

\noindent \textbf{{Robotics Manipulation}}. Table~\ref{tab:agentic_evaluation} shows that the \magma model without domain-specific finetuning already outperforms the recently proposed OpenVLA model pretrained for 27 epochs on the same amount of OXE data. Below, we testify the effectiveness of the finetuned \magma model by comparing it with OpenVLA in three settings:
\begin{itemize}
    \item Finetune on real robot data to evaluate on out-of-distribution manipulation tasks;
    \item Finetune in simulated robot settings with a limited number of trajectories using the LIBERO benchmark to evaluate \magma's capability of task adaptation; and
    \item Evaluate on the physical WidoxW 250 Arm.
\end{itemize}

\begin{figure}[!t]
    \centering
    \includegraphics[width=0.95\columnwidth]{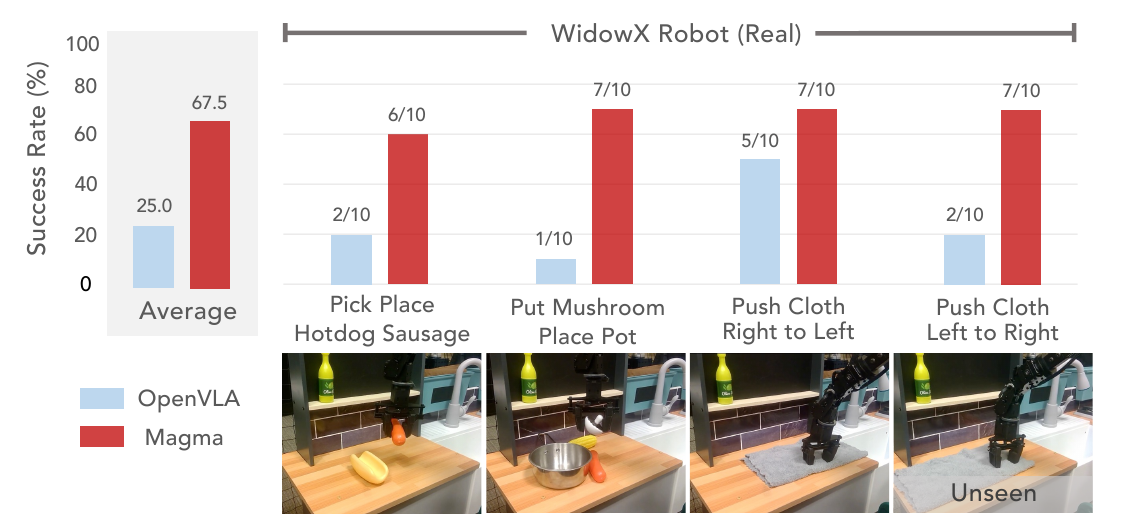}
    \vspace{-5pt}
    \caption{\textbf{Few-shot finetuning and generalization performance on real robot.} On a WidowX robot, we evaluate \magma on 4 tasks including diverse everyday object manipulation.}
    \label{fig:real}
\vspace{-10pt}
\end{figure}

\begin{figure}[!t]
    \centering
    \includegraphics[width=0.8\columnwidth]{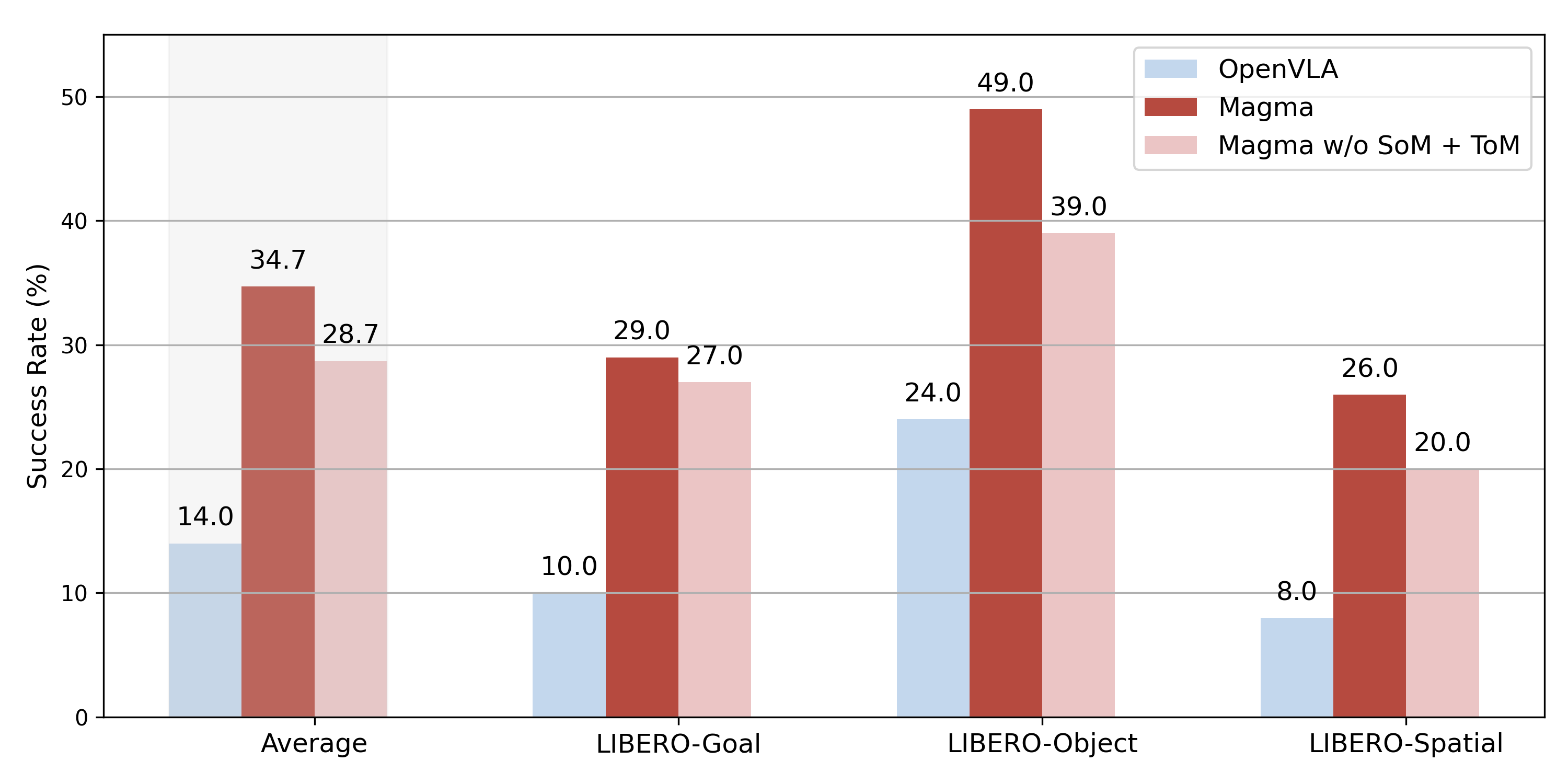}
    \vspace{-5pt}
    \caption{\textbf{Few-shot finetuning results on the LIBERO} simulation benchmark, using 10 trajectories per task for fine-tuning.}
    \label{fig:libero}
\vspace{-10pt}
\end{figure}


We collect four manipulation tasks each of which has roughly 50 trajectories (See details in our supplementary material), and finetune both OpenVLA and \magma jointly on these tasks. For evaluation, we perform 10 trials per task, ensuring the same initial states (positions and orientations of end-effector and objects) across models. As shown in Fig.~\ref{fig:real}, the results clearly demonstrate \magma's superior performance. For those challenging tasks that involve everyday objects like ``Pick Place Hotdog Sausage'', ``Put Mushroom in Pot'', and ``Push Cloth Right to Left'', OpenVLA can hardly accomplish the tasks, mainly because of the imprecise arm movement and object localization per our observation. In contrast, \magma performs well on these sophisticated tasks, largely owing to its strong spatial understanding and grounding capability obtained from pertaining. Additionally, we evaluate models' performance on an unseen task ``Push Cloth Left to Right'' which are not included in our finetuning dataset. \magma substantially outperforms the baseline, indicating a stronger ability to preserve pretrained knowledge and generalize to new tasks.

The efficient adaptation (via finetuning) capability of \magma is further validated through few-shot finetuning evaluations on the LIBERO benchmark. For each task suite in the benchmark, we sample only 10 trajectories for finetuning. During the evaluation, we perform 100 trials per task suite. The results, shown in Fig.~\ref{fig:libero}, indicate that \magma achieves a significantly higher average success rate in all task suites. 
Additionally, removing SoM and ToM during pretraining has a negative impact on model performance, underscoring the effectiveness of our pretraining method.

\begin{table}[t]
    \centering
    \resizebox{1.0\linewidth}{!}{
    \begin{tabular}{l|cccccccc}
    & \multirow{2}{*}{VSR} & \multirow{2}{*}{BLINK-val} & \multicolumn{3}{c}{SpatialEval\footnotemark } \\
    Model & & & \makecell{Spatial Map } & \makecell{Maze Nav.} & \makecell{Spatial Grid } \\
    \midrule
    GPT-4o & 74.8 & 60.0 & - & - & -\\
    Gemini  & - & 61.4 & - & - & -\\
          \hline
          \hline
    LLaVA-1.5-7B  &  57.1* & 37.1 & 28.4 & 28.8 &  41.6\\
    LLaVA-1.6-7B~\cite{liu2024llavanext}  & 52.2* & - & 28.0 & 34.8 & 32.2\\
    Qwen-VL-9.6B~\cite{Qwen-VL}  & - & 40.3 & 28.7 & 31.8 & 25.7\\
    \midrule
    \magma-8B~(Act\textsuperscript{w/o}) & 62.8 & 30.1 & 36.9 & \textbf{44.8} & 37.5    \\
    \magma-8B~(Full\textsuperscript{w/o}) & 58.1  & 38.3 & 27.5 & 33.5 & 47.3 \\
    \magma-8B~(Full) & \textbf{65.1} & \textbf{41.0} & \textbf{43.4} & 36.5 & \textbf{64.5} \\    
    \end{tabular}}
    \vspace{-3pt}
    \caption{\textbf{Spatial reasoning evaluations.} We use * to denote results that are obtained by us evaluating the provided model weights. Superscript `w/o' means models pretrained without SoM/ToM. 
    }
    \label{tab:spatial_eval_results}
\end{table}

\begin{figure}[t]
    \centering
    \includegraphics[width=1.0\columnwidth]{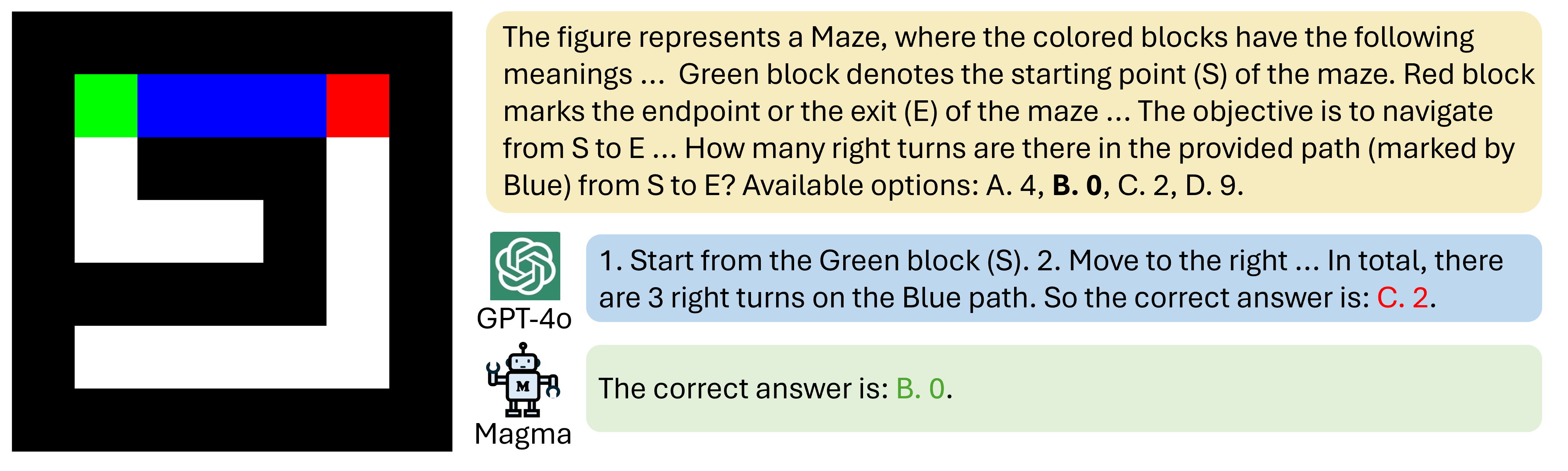}
    \vspace{-0.6cm}
    \caption{\textbf{Spatial evaluation predictions.} Spatial reasoning questions are challenging even for GPT-4o but \magma can answer relatively well despite relying on much fewer pretraining data.}
    \label{fig:spatial_eval_examples}
\end{figure}

\subsection{Evaluating Spatial Reasoning}
We attribute the much improved performance of our \magma model on the tasks of UI navigation and robotic manipulation, as shown above, to its improved ability to perform spatial reasoning. To verify this hypothesis, we evaluate the effectiveness of the spatial intelligence that is learned in our pretrained model on the challenging Visual Spatial Reasoning (VSR)~\cite{liu2023visual}, BLINK~\cite{fu2024blink} and SpatialEval~\cite{wang2024picture} benchmarks under the zero-shot setting. The results are summarized in Table~\ref{tab:spatial_eval_results}. We see that \magma outperforms existing approaches by significant margins on VSR and SpatialEval, and that \magma performs on par with CogVLM, despite only using $\sim$29M images for pretraining as compared to $\sim$1.5B images in the latter. In addition, our ablation study demonstrates the effectiveness of the SoM and ToM pretraining tasks in helping \magma improve its spatial reasoning capabilities. Last but not least, we also note the benefits of using video data during pretraining by showing that removing vidoes from training data leads to $\sim$8\% performance drop on BLINK. Finally, we also provide some example predictions of our \magma model in Figure~\ref{fig:spatial_eval_examples}. We observe that spatial reasoning questions are also challenging for SOTA proprietary models such as GPT-4o. Despite the lack of pretraining on data with mazes, we see that \magma is still able to answer spatial reasoning questions about them. 

\begin{table}[]
    \centering
    \resizebox{1.0\linewidth}{!}{
    \begin{tabular}{l|cccccccc}
    Model & VQAv2 & GQA & MME & POPE & TextVQA & ChartQA & DocVQA \\
    \midrule
    LLaVA-1.5-7B~\cite{li2023llava} & 76.6 & 62.6 & 1510.8 & 85.9 & 46.1 & 18.2 & 28.1 \\
    LLaVA-Next-7B~\cite{liu2024llavanext} & 80.1 & \textbf{64.2} & 1519.3 & \textbf{86.4} & 64.9 & 54.8 & 74.4 \\
    \midrule
    \magma-8B~(SFT) &79.5 & 61.5 & 1510.1 & 86.2 & 67.7 & 73.0 & 80.4 \\    
    \magma-8B~(Act\textsuperscript{w/o}) &81.3 & 63.5 & 1559.5 & 86.1 & 69.8 & 71.0  & 84.1 \\ 
    \magma-8B~(Full\textsuperscript{w/o}) & 81.3& 62.9 & 1576.0 & 86.3 & 69.6& 71.7 & 83.8 \\ 
    \magma-8B~(Full) & \textbf{81.4} & 64.0 & \textbf{1588.7} & 86.3 & \textbf{70.2} & \textbf{76.2} & \textbf{84.8} \\
    \end{tabular}}
    \vspace{-5pt}
    \caption{\textbf{Finetuned performance on multimodal image understanding tasks}. Pretraining on full set with SoM and ToM~(last row) attains the overall best performance compared with our own baselines and counterparts of the same model class.
    }
    \vspace{-1.5em}
    \label{tab:image_qa_results}
\end{table}

\footnotetext{We evaluate our model using the standard option matching before the official evaluation pipeline was released and will update in the next version.}

\begin{table*}[t]
    \centering
    \small
    \resizebox{\textwidth}{!}{%
    \begin{tabular}{l|lcccccccccccccc}
        \multirow{2}{*}{Method} &  \multirow{2}{*}{Backbone} & \multicolumn{1}{c}{IntentQA}  & \multicolumn{1}{c}{Next-QA}  & \multicolumn{3}{c}{VideoMME (w/o subs)} & \multicolumn{4}{c}{MVBench} \\
         &   & Overall & Overall & Short & Medium & Long & Action Prediction & Action Sequence & Action localization  & Overall\\        
        \toprule
        \multirow{1}{*}{Gemini-1.5~\cite{geminiteam2024gemini}} & - & - & - & 81.7 & 74.3 & 67.4 & - & - & - & 37.7\\
         \multirow{1}{*}{GPT-4V~\cite{achiam2023gpt}} & GPT-4  & - & - & 70.5 & 55.8 & 53.5  & - & - & - & 43.7\\      
        \hline
        \hline
         LLaVA-OV~\cite{li2024llavaonevision} & Qwen2-7B   & - & \underline{79.4} & \underline{68.1} & \underline{54.9} & \textbf{47.8} & 46.0 & 74.5 & 48.0 & 56.7\\
          Long-Llava 9B~\cite{wang2024longllava}  & Long-Llava 9B & - & - & 52.4 & 42.2 & 36.4 & - &- &- & 49.1\\
           LongVA~\cite{zhang2024long} & Qwen2-7B  & - & 69.3 & 61.1 & 50.4 & \underline{46.2} & 49.0 & 53.0 &42.5 & 51.3 \\

          ShareGPT4Video~\cite{chen2024sharegpt4video} & LLaMA3-8B & - & - & 48.3 & 36.3 & 35.0 & 40.0 & 49.5 & 41.5 & 51.2\\

          Video-Llama2~\cite{cheng2024videollama} & Llama2-7B  & - & - & 55.9 & 45.4 & 42.1 & - &- &- & 34.1\\
          
          Video-Chat2~\cite{li2024mvbenchcomprehensivemultimodalvideo} & Mistral 7B  & - & 43.3 & 48.3 & 37.0 & 33.2 & 47.5 & \underline{75.0} & \underline{50.5} & \textbf{60.4}\\
          Video-Llava~\cite{lin2023video} & Vicuna-7B  & - & 51.4 & 45.3 & 38.0 & 36.2 & \underline{50.0} & 38.5 & 30.5 & 43.0\\ 
          IG-VLM~\cite{kim2024image} & Vicuna-7B & \underline{60.3} & - & - & - & - & - & - & - & -\\
          SF-LLaVA~\cite{xu2024slowfast} & Vicuna-7B & 60.1 & - & - & - & - & - & - & - & -\\ 
          \midrule
           Magma-8B~(Ours) & LLaMA3-8B  & \textbf{88.6} & \textbf{80.9} & \textbf{72.9} & \textbf{55.8} & 44.3 & \textbf{65.0} & \textbf{79.0} & \textbf{55.5} & \underline{59.4}\\
    \end{tabular}}
    \caption{\textbf{Zero-shot Video QA benchmarks.} We compare our Magma model to other state-of-the-art approaches with comparable numbers of parameters. Our \magma model performs competitively and even outperforms some state-of-the-art approaches such as Video-Llama2 and ShareGPT4Video on most benchmarks, despite using much fewer video instruction tuning data.}
    \label{tab:video_qa_results}
\end{table*}

\subsection{Evaluating Multimodal Understanding}

\paragraph{Image instruction tuning.} To further assess \magma's multimodal understanding capability, we conduct continuous finetuning on our Magma-SFT-820K data. Then, we compare the finetuned \magma model with existing VLMs on a suite of commonly used image reasoning benchmarks, \eg MME and GQA. As shown in Table~\ref{tab:image_qa_results}, \magma outperforms recently-proposed VLMs on most of the tasks, with notable gains of $\sim$5\% and $\sim$22\% on TextVQA and ChartQA, respectively. Similarly to our observations in Table~\ref{tab:spatial_eval_results}, our ablation study highlights the effectiveness of using SoM and ToM for pre-training, which leads to $\sim$ 5\% improvement in ChartQA.

\vspace{-0.3em}
\paragraph{Video Instruction Tuning} 
In Table~\ref{tab:video_qa_results}, we report the performance of our \magma model on multiple challenging video question answering (QA) benchmarks including IntentQA~\cite{Li2023IntentQACV}, NextQA~\cite{xiao2021next}, VideoMME~\cite{fu2024videommefirstevercomprehensiveevaluation} and MVBench~\cite{li2024mvbenchcomprehensivemultimodalvideo}. We use the LMMs-Eval framework~\cite{lmms_eval2024} for the latter three benchmarks to ensure reproducibility of our evaluation results. 

The results demonstrate the effectiveness of our pretraining approach, where we outperform most state-of-the-art models with comparable number of parameters consistently across the different benchmarks. For instance, our \magma model achieves a performance gain over the IG-VLM and SF-LLaVA models by approximately 28\%. The IntentQA benchmark evaluates a model's capability to discern the intentions behind observed actions in videos. Thus, the significant improvement on this dataset achieved by \magma can possibly be attributed to the effectiveness of our ToM pretraining task, where it encourages the model to reason about temporal dynamics in future video frames. This is also corroborated by the notable improvement on the subtask of action prediction in MVBench that \magma obtains over state-of-the-art models such as VideoChat2 and LLaVA-OV.

State-of-the-art video LMMs often rely on much large video and text datasets such as Webvid and ShareGPT4Video for pretraining and these datasets span over 4M samples with curated text. Moreover, the aforementioned models also use a higher number of frames during pretraining. In contrast, even when multi-frame pretraining is performed in our case, we only use a maximum of 4 frames due to computational constraints. Thus, it is especially significant that \magma outperforms approaches such as LLaVA-OV and ShareGPT4Video on VideoMME and MVBench, since these approaches often use larger instruction tuning datasets that include both image and video data. Additionally, as evidenced by the performance gain obtained by \magma over the proprietary GPT-4V model, we note that such improvements in results are not solely due to using a more recent and powerful language model like LLama-3. It is also notable that Magma achieves substantially better performance than LongVA, despite using only 32 frames instead of the 64 frames used by the latter. 
\section{Conclusion}
We present the \magma foundation model that can understand and act on multimodal inputs to complete agentic tasks in different environments. 
Our experiments show that the use of SoM and ToM prediction tasks in pretraining helps the model learn to ground and plan actions, respectively. In our experiments, \magma shows strong spatial-temporal reasoning ability and significantly outperforms baselines on downstream UI navigation and robotic manipulation tasks.

\paragraph{Social Impacts and Limitations.}

To develop a foundation model with both verbal and spatial intelligence capable of handling diverse agentic tasks in digital and physical environments, we curated a comprehensive pretraining dataset from a wide range of image, video, and robotics domains: 
\begin{itemize}
    \item \textbf{UI navigation data}. We leverage two pretraining datasets SeeClick and Vision2UI.
    \item \textbf{Instructional videos}. As our goal was to learn an agentic model that can undertake daily tasks like humans, we compile the videos from Epic Kitchen, Ego4d, Something-Something v2 and other instructional videos.

    \item \textbf{Robotics manipulation data}. For robotics task, we follow OpenVLA to leverage the robotics data in Open-X-Embodiment.
    \item \textbf{Multimodal understanding data}. Lastly, we include a small set of multi modal pretraining data ShareGPT4V, and instruction tuning data LlaVA-1.5 plus a number of other domain-specific data to retain the generic multimodal understanding capability of the pre-trained model. 
\end{itemize}
 
The data markup of the robotics and UI navigation data is fairly standardized focusing on generic manipulation tasks (“Place x object on y object”) and generic UI navigation tasks (“Click search button”). We, however, performed a detailed data reflection exercise on the video data of people performing certain tasks. The core inferences we took from these videos were the trajectory of objects over time when the tasks were performed.  

We note that the distribution of identities and activities in the instructional videos are not representative of the global human population and the diversity in society. We are cognizant of the unintended societal, gender, racial and other biases in training with these data, so we will ensure required disclaimers are in place when publishing the models. The training dataset, task list and descriptions focus on the next action to perform only – not describe, act on, or perform any analysis on the subject itself. While there can be unintended outputs from the model based on adverse task descriptions, we will ensure to highlight the use cases the model was trained for and it’s intended use.  

\paragraph{\textbf{Responsible AI}.} It is important to note that the model is specifically designed for UI navigation in a controlled Web UI and Android simulator, and robotic manipulation tasks and should not be broadly applied to other tasks. The recommended usage is within the settings they were trained on, namely, an enclosure equipped with a robotic arm and everyday objects for robotic manipulation and an android simulator running on a computer for UI manipulation. For UI navigation task, researchers should make sure that a human is in the loop and in control for every action the agentic system generates. Since the model cannot act by itself, the sub-module a researcher uses to actually perform the UI navigation action should ensure that no unintended consequences can occur as a result of performing the UI action proposed by the model. 

The model by itself demonstrates good-enough capability in UI navigation and robotic manipulation, but is not usable as is for exploitation scenarios. A threat actor, can however use specific training data for a specific malicious task, to leverage the model as a base to perform automated UI navigation. This is a generic risk associated with the agentic models. 

\paragraph{Acknowledgments.} We would also like to thank Professor Yong Jae Lee for thoughtful discussions, Xiyang Dai for valuable discussions and data support, Mei Yang and Denny Sun for early data engineering effort, and Swadheen Shukla for internal RAI and data reviews. We would also like to thank Doug Burger and Desney Tan for the multifaceted leadership support.

{
    \small
    \bibliographystyle{ieeenat_fullname}
    \bibliography{main}
}
\clearpage
\appendix
\definecolor{custom_blue}{RGB}{235,244,253}
\tcbset{
  aibox/.style={
    width=\textwidth,
    top=10pt,
    colback=white,
    colframe=black,
    colbacktitle=black,
    center,
  }
}

\newtcolorbox{graybox}[1][]{colback=lightgray!20, colframe=lightgray!50!black, boxrule=0pt, enhanced, #1}
\newtcolorbox{redbox}[1][]{colback=red!10, colframe=red!50!black, boxrule=0pt, enhanced, #1}
\newtcolorbox{greenbox}[1][]{colback=green!10, colframe=green!50!black, boxrule=0pt, enhanced, #1}
\newtcolorbox{magentabox}[1][]{colback=magenta!10, colframe=magenta!50!black, boxrule=0pt, enhanced, #1}
\definecolor{lightgreen}{RGB}{144, 238, 144} 

\newtcolorbox{AIbox}[2][]{aibox,title=#2,#1}
\tcbset{
  aiboxsmall/.style={
    width=0.62\textwidth,
    top=10pt,
    colback=white,
    colframe=black,
    colbacktitle=black,
    enhanced,
    top,
    attach boxed title to top left={yshift=-0.1in,xshift=0.15in},
    boxed title style={boxrule=0pt,colframe=white,},
  }
}   
\definecolor{commentcolor}{RGB}{34,139,34} 
\newcommand{\myalgorithm}{
\begingroup
\removelatexerror
\begin{algorithm*}[H]
      \caption{\modelname PyTorch pseudocode.}
      \label{alg:pseudocode}
      \scriptsize
          \Comment{
          $\mathbf{H}_0$: Input embeddings for LLM (Original inputs args for traditional LMM); \\
          $vis\_pos$: the location of visual tokens; \\
          $\mathbf{X}$, $\mathbf{X^{stack}}$: Original visual tokens, Extra high-resolution visual token list; \\
          $l_{start}$, $n$: Index of starting layer, and layer interval for stacking.
          }
  \Function{forward($\mathbf{H}_0$, $\mathbf{X^{stack}}$, $l_{start}$, $n$, $vis\_pos$)}{
     \var{$\mathbf{H}$ = $\mathbf{H}_0$}
     
     \For{($idx$, \var{TransformerLayer)} in enumerate(\var{self.layers})}{
        \Comment{\modelname:}
        \If{$idx$ >= $l_{start}$ \& $(idx - l_{start}) \% n == 0$}
        {
            $\mathbf{H}[vis\_pos]$ += $\mathbf{X^{stack}}[(idx - l_{start})//n]$
        }

        \Comment{Original Transformer:}
        $\mathbf{H}$ = \var{TransformerLayer}($\mathbf{H}$)
     }
  }
\end{algorithm*}
\endgroup}

\maketitlesupplementary

\section{Pretraining and Finetuning}

\begin{table}[h]
    \centering
    \resizebox{0.99\linewidth}{!}{
    \footnotesize
    \begin{tabular}{l | c c c c}
        Setting & Pretraining & \multicolumn{3}{c}{Finetuning}
         \\
        & & UI & Image/Video & Real Robot\\
        \midrule
        batch size          &  1024 & 32   \\
        base learning rate  &  1e-5 & 1e-5  & 1e-5 & 1e-5 \\
        learning rate scheduler & Constant & Cosine & Cosine & Constant \\
        training epochs     &  3  & 3 & 1 & 20 \\
        optimizer          & adamw   & adamw & adamw   & adamw 
        \\
        \midrule
        Image Resolution & 512 & 768 & 768 & 256 \\
        Number of Crops & 4 or 1  & 4 & 4 or 1 & 1\\
        
    \end{tabular}}
    \caption{Experimental settings pretraining and finetuning of \magma models. We maximally use either 32 Nvidia H100s or 64 AMD MI300 GPUs for all training jobs.}
    \label{tab:settings_pt_ft}
\end{table}

For all the model variants, we use the same training recipe as shown in Table~\ref{tab:settings_pt_ft}. To handle different image resolutions from different datasets, we also use a multi-crop strategy to enable batch forward for a given minibatch, though the ConvNext vision backbone can naturally support arbitrary resolutions. Specifically, for our pretraining, we use 512 as the base image size, and resize an input image maximally to 4 crops for UI and image pretraining data, while use 1 crop for video and robotics data.

For downstream finetuning, we following common practice to tune the pretrained magma model as shown in Table~\ref{tab:settings_pt_ft} right. As mentioned above, the vision encoder can be effortlessly adapted to different image resolutions required for different tasks.
\section{Datasets}
\label{sec:supp_pretraining_data}

\begin{table}[t]
    \centering
    \resizebox{0.9\linewidth}{!}{
    \begin{tabular}{lcc}
     Source & Task & Size \\
     \toprule
     \multirow{4}{*}{SeeClick-Web} & text\_2\_point  & 271K \\
     & text\_2\_bbox & 54K \\
     & point\_2\_text & 54K \\
     & bbox\_2\_text & 54K \\
      \midrule
     \multirow{4}{*}{SeeClick-Mobile} & text\_2\_point  & 274K \\
     & text\_2\_bbox & 56K \\
     & UI summarization & 48K \\
     & widget captioning & 42K \\
      \midrule
     \multirow{4}{*}{Visison2UI} 
     & input\_2\_point & 980K\\
     & input\_2\_bbox & 982K \\
     & text\_2\_point  & \textcolor{Gray}{794K} \\
     & text\_2\_bbox & \textcolor{Gray}{774K} \\
     & point\_2\_text & \textcolor{Gray}{199K} \\
     & bbox\_2\_text & \textcolor{Gray}{193K} \\
     \bottomrule
     \rowcolor{custom_blue}  Magma-PT-UI~(Ours) & Mixed & 2.8M \\
    \end{tabular}}
    \vspace{-5pt}
    \caption{Statistics of UI related pretraining data.}
    \label{tab:ui_pretrain}
\end{table}

\subsection{Pretraining Data}
Due to space constraints, we briefly introduced the datasets for our pretraining in Sec~4.1 of our main submission. To ensure the reproducibility of our pretraining stage, we provide additional details of our pretraining data below.

\begin{figure*}[t]
    \centering
    \includegraphics[width=2.\columnwidth]{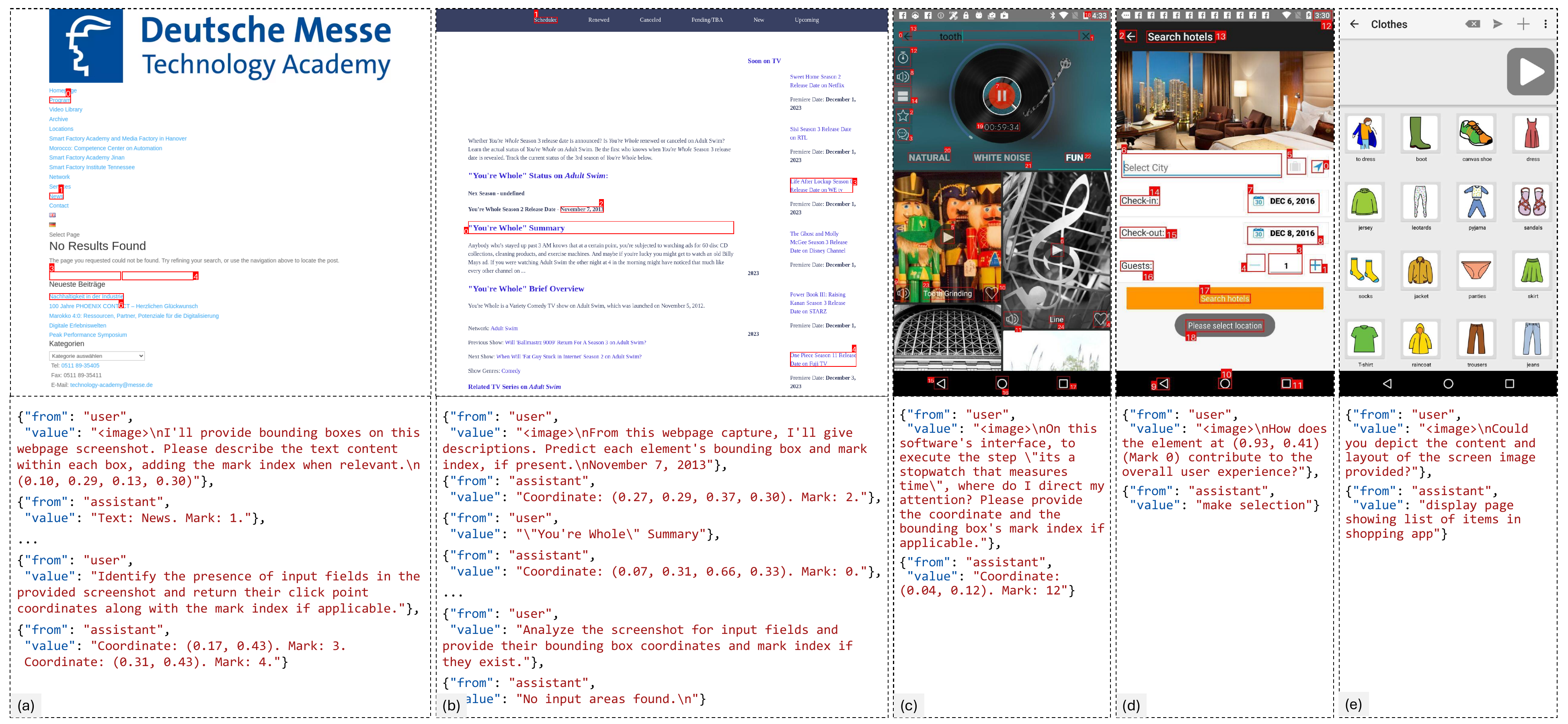}
    \vspace{-5pt}
    \caption{\textbf{Training samples in our Magma-PT-UI.} It covers a wide range of action grounding and UI understanding tasks including: (a) Given the bounding box or point coordinates as the query, assistant should return the natural language description or the content. (b) Given the natural language or the exact content as the query, assistant should return the value of the bounding box coordinates.. (c) Given the natural language as the query, assistant should return the value of the point coordinate. (d) Widget captioning. (e) UI summarization.}
    \label{fig:magma_pt_ui}
\vspace{-10pt}
\end{figure*}

\subsubsection{UI Navigation} 
Our pretraining data related to UI agent are sourced from two datasets, SeeClick~\cite{seeclick} and Vision2UI~\cite{gui2024vision2uirealworlddatasetlayout}. We further process these source data by adding marks on screenshots to provide grounded supervisions.

\noindent\textbf{SeeClick}. We generally follow the original procedure and make the following modifications to associate with the Set of Mark~\cite{yang2023set} strategy.
For each webpage screenshot, multiple (text, bounding\_box) pairs are available. Therefore, we directly overlay all the bounding boxes with corresponding marks on the screenshot.
For each mobile screenshot, only a single (text, bounding\_box) pair is available in the SeeClick data. To enrich the pairs, we incorporate additional pairs from the RICO dataset~\cite{deka2017rico}, and employ an OCR tool to obtain text boxes. Finally, we display the enriched bounding boxes along with their corresponding marks on the mobile screenshot.

\noindent\textbf{Vision2UI}. We consider all bounding boxes whose ``content'' property is not null. To prevent the marks from overwhelming the main content of the webpage, we sample bounding boxes with varying probabilities based on their "type" property. Specifically, we assign a sampling weight of 0.5 to boxes of type \texttt{h1}, \texttt{h2}, \texttt{a}, \texttt{button}, \texttt{option}, and \texttt{nav} with 0.5, while other types are weighted at 0.1. Given the high importance of input areas for interaction, we include boxes of type \texttt{input} directly without sampling for mark plotting. After obtaining the elements of high interest, we apply similar tasks as SeeClick~\cite{seeclick} to produce the instruction data, including (a) grounding task, which involves two forms: predicting center point coordinates (text\_2\_point) and predicting bounding box (text\_2\_bbox); (b) generating text for elements, categorized into predicting text based on the coordinates of center points (point\_2\_text) or bounding boxes (bbox\_2\_text); and further introduce the task of (C) locating input fields, including predicting center point coordinates (input\_2\_point) and bounding box coordinates (input\_2\_bbox) of the input fields.

Given a webpage, since the first two categories of tasks are grounding or generating texts for the same group of web elements, we further weight the four subtasks, \textit{i.e.}, (text\_2\_point), (text\_2\_bbox), (point\_2\_text), and (bbox\_2\_text) with [0.4, 0.4, 0.1, 0.1], and sample only one of them to construct the pretraining data. Similarly, we sample one subtask from (input\_2\_point) and (input\_2\_bbox) with equal probabilities. We merge the sampled subtasks from the same webpage into one example to improve training efficiency. We denote the full pretraining data related to UI by Magma-PT-UI, and list the sizes of individual subsets in Table~\ref{tab:ui_pretrain}.

\begin{figure*}
    \centering
    \includegraphics[width=0.95\linewidth]{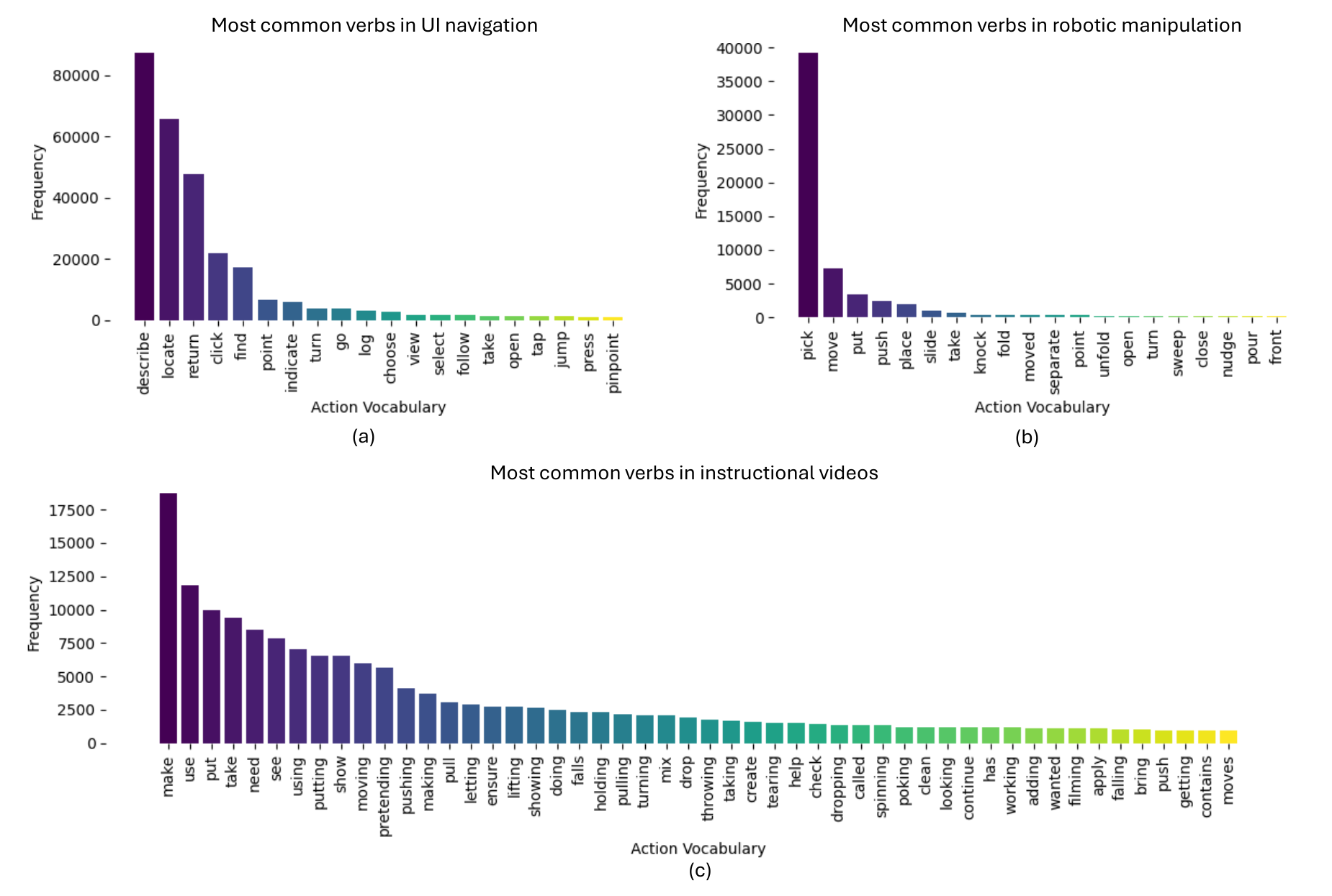}    
    \hfill
    
    \caption{\textbf{Action distributions in three types of action-oriented pretraining datasets}. (a) UI Navigation; (b) Robotic Manipulation; (c) Instructional Videos.}
    \label{fig:data_dist}
\end{figure*}
\subsubsection{Instructional Videos} 
As mentioned in the main submission, we curate the supervisions from human instructional videos to learn the agentic capability for our model. To cover different scenarios, we considered both 3rd point view videos and egocentric videos. In particular, we start with Epic-Kitchen~\cite{Damen2018EPICKITCHENS} video data sets considering that their text annotations are relatively high quality. Afterwards, we expand to Something-Soomething v2~\cite{materzynska2020something} to include more human-object interactions, and Ego4D~\cite{grauman2022ego4dworld3000hours} and other related instructional videos for scaling up.

\noindent \textbf{Epic-Kitchen}~\cite{Damen2018EPICKITCHENS}. Epic-Kitchen contains 495 egocentric videos recorded by 32 participants in kitchen rooms. Each video contains a number of segments labeled with narrations, start and end frame ids. However, the original video narrations (\eg, ``open door'') are too coarse to depict the human actions in a certain time frame.

For the videos in Epic-Kitchen, we apply the video preprocessing method as discussed in Sec~4.2 of our main submission. Concretely, for each of the original video segments in the dataset, we run PySceneDetect to detect the temporal boundaries and split them into sub-segments. During our model pretraining, the textual annotations are used in two ways. Our model is asked to predict the detailed description in the first frame.  In addition, they are used as the task description as input to the model for predicting the traces of marks.

\noindent \textbf{Sth-Sth-v2}~\cite{materzynska2020something}, \textbf{Ego4D}~\cite{grauman2022ego4dworld3000hours}. 
The Sth-Sth v2 dataset is a comprehensive collection of labeled video clips featuring humans performing predefined actions with everyday objects. The list of action classes spans a wide variety of atomic actions, including but not limited to ``pushing something from right to left'', ``throwing something'' and ``covering something with something''. In total, the dataset contains 220,847 seconds-long video clips. To create our pretraining data, we only leverage the videos in the train and validation splits. This amounts to around 160K video clips. We note that we do not use PySceneDetect for Sth-Sth v2 since the original video clips have been highly curated.

The Ego4D dataset is a large-scale egocentric dataset that contains approximately 3,025 hours of videos. It comprises over 3,670 hours of video footage captured from wearable cameras across a diverse environments and activities. The dataset spans a wide range of real-world scenarios, including daily activities and social interactions. Given the duration of these videos can span over 30 minutes, we leverage the original dense caption annotations that are provided to split each videos into seconds-long segments with consistent views.

\subsubsection{Robotic Manipulation} 
\begin{figure*}
    \includegraphics[width=1.0\linewidth]{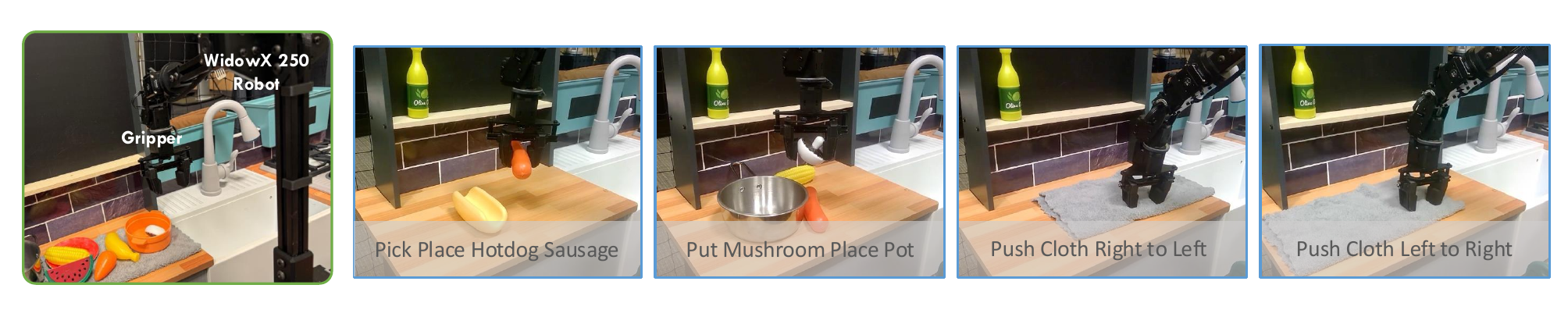}
    \vspace{-1.5em}
    \caption{\textbf{Real robot setup.} Magma is deployed on a WidowX 250 robot arm to perform a sequence of kitchen manipulation tasks including object pick-place and soft manipulation.}
    \label{fig:supp_real}
\vspace{-3pt}
\end{figure*}
We follow the training recipe in OpenVLA~\cite{kim2024openvla} to prepare our pretraining data for robotics manipulation. Specifically, we take the data mixture ``siglip-224px+mx-oxe-magic-soup'' as in OpenVLA, which gives us 9.4M image-language-action triplets, extracted from 326K trajectories, from 23 separate datasets. 

\subsubsection{Multimodal Image Understanding} 
We simply include the 1.2M synthetic image-text pairs in ShareGPT4V~\cite{chen2023sharegpt4v} and 665K image instruction tuning data collected by LLaVA-1.5~\cite{liu2024llavanext} as our multimodal image pretraining data. The former helps our pretrained model to have a global understanding of visual contents, while the latter helps to get the model familiar with various types of human instructions. We denote this dataset by Magma-PT-Image.

\subsubsection{Data Statistics}
Given our goal of training a general vision-language-action foundation model, we analyze the distribution of verbs present in the text annotations of the UI and robotic manipulation as well as instructional video datasets in Figure~\ref{fig:data_dist}. We see that the text annotations in the UI navigation component contain many helpful verbs that help guide agents to achieve a specific task such as ``locate'' and ``turn''. This is complemented by the more action-oriented words in the vocabulary of the robot manipulation component, including ``pick'', ``push'' and ``slide''. Such annotations are especially valuable in helping our \magma model to learn to reason about interactions with everyday objects. Finally, we also scale up the amount of training data and diversity of verbs by including data from instructional videos (Figure~\ref{fig:data_dist}c). As evidenced by the relatively high frequency of words such as ``lifting'' and ``throwing'', such annotations can be very beneficial for gaining a stronger understanding the of temporal dynamics involved in common activities. More importantly, the diversity of activities present in these datasets can be effective at helping the model generalize better to a larger variety of tasks.
\subsection{Downstream Data}

\subsubsection{UI Agent Navigation}

We evaluated the UI grounding and navigation capability mainly on three datasets, ScreenSpot~\cite{seeclick}, Mind2Web~\cite{mind2web} and AITW~\cite{aitw}.

\noindent \textbf{ScreenSpot} is a benchmark used to evaluate the UI action grounding proposed in~\cite{seeclick}. It consists of 600 screenshots images associated with 1.2K instructions spanning iOS, Android, macOS, Windows,
 and web pages. The evalaution covers both text
based elements and a variety of widgets and icons. To evaluate the zero-shot action grounding performance for our model, we use OmniParser~\cite{lu2024omniparser} to help parse the screenshot and propose actionable regions/icons/buttons. We used the sample code and default settings provided in the official repo. For these candidate regions, we overlay numeric marks and ask our model to pick one.

\noindent\textbf{Mind2Web} is first proposed in~\cite{mind2web} for text-based web agent. For fair comparison among vision-based web agent, we follow the protocol proposed in SeeClick~\cite{seeclick}. Given a webpage, we convert it into a screenshot associated with ground-truth bounding boxes to which the actions should be applied. As the original screenshot of the full website is usually out of the scope of display. We follow a similar way as in~\cite{seeclick} to crop the region of interests centering around the ground truth boxes, which gives us a local screenshot as wide as original webpage but with maximal height 1344. To propose the candidate marks for our model, we directly exploit the candidate ranks provided in Mind2Web, and use the top 30 candidates for evaluation.

\noindent \textbf{AITW} is a dataset originally collected in~\cite{aitw} for navigation of the android UI. The original dataset contains up to 715K trajectories, resulting in 5.7M screenshots. In our experiments, to examine the efficient finetuning performance, we alteratively follow the same protocol in SeeClick~\cite{seeclick} and include a much smaller number of training samples. Specifically, there are 545, 688, 306, 700, 700 instructions from General/Install/GoogleApps/Single/WebShopping, respectively. $80\%$ of each split is used for training and the remainder is used for evaluation. Instead of finetuning our model for each category, we jointly finetune our pretrained \magma on the combined data and evaluate across all categories using a single model.

\subsubsection{Robot Manipulation}

\noindent\textbf{Simulator}. We employ SimplerEnv~\cite{li24simpler} as the main testbed for our learned robot policy. As we do not need to tune our model on the simulated trajectories, we simply report the numbers following the protocol proposed in the original work.

\noindent\textbf{Real-world Setting}. We design four tabletop manipulation tasks for our physical WidowX-250 robot setup as shown in \ref{fig:supp_real}. As with BridgeData-v2, the RGB image observations from the robot are captured using a stationary third-person camera, maintaining a resolution of $256 \times 256$. For finetuning our pretrained \magma model, we collect approximately 50 robot demonstration trajectories for each task as our finetuning dataset. Our experimental design includes classic soft object manipulation and pick-and-place operations tasks. Detailed language instructions for the designed tasks are presented below. For each trial, we randomize the initial location of the target object and include 2-3 random distracting objects (e.g., corn, eggplant) in the scene. For reproducibility, we release the collected robot trajectories.

\noindent Tasks included in the finetuning dataset:
\begin{itemize}
\item \textbf{Hot dog assembly}: Pick up the hot dog sausage from the desk and place it into the bun. The trial is counted as success only when the robot successfully grasps the sausage and accurately places it within the hot dog bun.

\item \textbf{Mushroom placement}: Pick up the mushroom and place it into the pot. The trial is counted as success only when the robot correctly grasps the mushroom and places it into the cooking pot without dropping or misaligning it.

\item \textbf{Cloth pushing}: Push the cloth from right to left across the surface. The trial is counted as success only when the robot successfully manipulates the cloth in the specified direction without disturbing other objects on the surface.
\end{itemize}

\noindent Unseen task for evaluating generalization:
\begin{itemize}
\item \textbf{Bidirectional cloth manipulation}: Push the cloth in both directions while maintaining its shape. This task examines the model's spatial understanding and reasoning capabilities, as it requires generalization from unidirectional pushing in the training data to bidirectional manipulation in novel scenarios.
\end{itemize}

\subsubsection{Image Instruction Tuning} 

We show a breakdown of our 820k Magma image instruction tuning data in Table~\ref{tab:magma_820k}. As the 760k image instruction tuning data used in LLaVA-1.6~\cite{liu2024llavanext} is not released, we follow their guidance to curate 748k public available data including ShareGPT~\cite{sharegpt}, LLaVA-Instruct~\cite{liu2023llava}, ShareGPT4V~\cite{chen2023sharegpt4v}, LAION-GPT4V~\cite{laion4v}, VQAv2~\cite{goyal2017making}, GQA~\cite{hudson2019gqa}, OKVQA~\cite{marino2019ok}, OCRVQA~\cite{mishra2019ocr}, ChartQA~\cite{masry2022chartqabenchmarkquestionanswering}, DVQA~\cite{kafle2018dvqa}, DocVQA~\cite{mathew2021docvqa}, AI2D~\cite{ai2d}, SynthDog-EN~\cite{kim2022ocr}, A-OKVQA~\cite{schwenk2022okvqa}, RefCOCO~\cite{kazemzadeh2014referitgame} and VG~\cite{krishna2017visual}. To complement the claimed ``improved reasoning, OCR and world knowledge'', we resort to a few other open-sourced datasets including InfoGraphicsVQA~\cite{mathew2021infographicvqa}, augmented ChartQA~\cite{masry2022chartqabenchmarkquestionanswering}, FigureQA~\cite{kahou2018figureqaannotatedfiguredataset}, TQA~\cite{tqa2017} and ScienceQA~\cite{lu2022learnscienceqa}. We denote the full set by Magma-SFT-Image.

\subsubsection{Video Instruction Tuning}
For comparisons with state-of-the-art video LMMs, we adopt the LLava-Video-178K dataset~\cite{zhang2024llavanextvideo} for instruction tuning. It consists of approximately 1.6M video and text instruction samples from 178K videos. The dataset is compiled from multiple video sources ranging from Charades~\cite{sigurdsson2016hollywood}, Sth-SthV2~\cite{materzynska2020something} to Kinetics-700~\cite{carreira2019short}. We refer interested readers to the original papers for more details.

\begin{table}[t]
    \centering
    \resizebox{1.0\linewidth}{!}{
    \begin{tabular}{l|cc}
     Dataset & Size & Domain \\
     \toprule
     ShareGPT~\cite{sharegpt}   & 40K  & Text \\
     ShareGPT4V~\cite{chen2023sharegpt4v} & 39K & General \\
     LLaVA-Instruct~\cite{liu2023llava} & 158K & General \\ 
     LAION-GPT4V~\cite{laion4v} & 11K & General \\     
     \hline
     VQAv2~\cite{goyal2017vqav2} & 83K & General VQA \\
     GQA~\cite{hudson2019gqa} & 72K & General VQA \\
     OKVQA~\cite{schwenk2022okvqa} & 9K & Knowledge VQA \\
     OCRVQA~\cite{mishra2019ocr} & 80K & OCR VQA \\
     ChartQA~\cite{masry2022chartqabenchmarkquestionanswering} & 7K & Chart VQA \\
     DVQA~\cite{kafle2018dvqa} & 16K & Chart VQA \\
     DocVQA~\cite{mathew2021docvqa} & 10K & Document VQA \\
     AI2D~\cite{ai2d} & 2K & Infographic VQA \\
     SynthDog-EN~\cite{kim2022ocr} & 20K & Document Understanding\\
     A-OKVQA & 66K & Knowledge VQA \\
     RefCOCO~\cite{yu2016modeling-refcoco} & 48K & Grounding Desc. \\
     VG~\cite{krishna2017visual} & 86K & Referring Exp. \\
      \midrule
     InfographicsVQA~\cite{mathew2021infographicvqa}    &  24k & Infographic VQA \\
     ChartQA~(Aug)~\cite{masry2022chartqabenchmarkquestionanswering} & 20k & Chart VQA \\
     FigureQA~\cite{kahou2018figureqaannotatedfiguredataset} & 20k & Chart/Figure VQA \\
     TQA~\cite{tqa2017} & 1.5k & Textbook VQA \\
     ScienceQA~\cite{lu2022learnscienceqa} & 5k & Textbook VQA \\
     \bottomrule
     \rowcolor{custom_blue} Magma-SFT-Image~(Ours)  & 820k & Mixed \\
    \end{tabular}}
    \vspace{-5pt}
    \caption{A detailed breakdown of our 820k Magma image instruction tuning data used in our multimodal image understanding experiments shown in Table~5 in our main submission.}
    \label{tab:magma_820k}
\end{table}

\subsubsection{Details about SoM for training and evaluation}

we exploit three ways to extract the candidate bounding boxes for the SoM prompt: 
\begin{itemize}
    \item \textbf{DoM Tree}. In addition to the bounding boxes extracted from HTML code~\cite{seeclick,gui2024vision2uirealworlddatasetlayout}, we further annotate the mobile screenshots in SeeClick data with bounding boxes derived from Android view hierarchies~\cite{rico_semantics}. These annotations are used during our model pretraining.
    \item \textbf{Vision model}. For zero-shot evaluation on Screenspot~\cite{seeclick}, we exploit the OmniParser model~\cite{lu2024omniparser} to make a fair comparison with the state-of-the-art methods~\cite{lu2024omniparser,seeclick}. Note that we only use the bounding boxes without local semantics. The original bounding boxes in AITW~\cite{aitw} are identified using an OCR model and IconNet~\cite{rico_semantics}. 

    \item \textbf{Language model}. For evaluation on As discussed earlier, we directly apply the predictions provided by Mind2Web~\cite{mind2web} using a pretrained language model DeBERTa-v3-base. This model gives approximately $85\%$ recall@50.
\end{itemize}

\label{sec:supp_techniques}

\section{Qualitative Analysis}
\label{sec:qualitative}

\subsection{UI Navigation}

\begin{figure*}
    \includegraphics[width=1.0\linewidth]{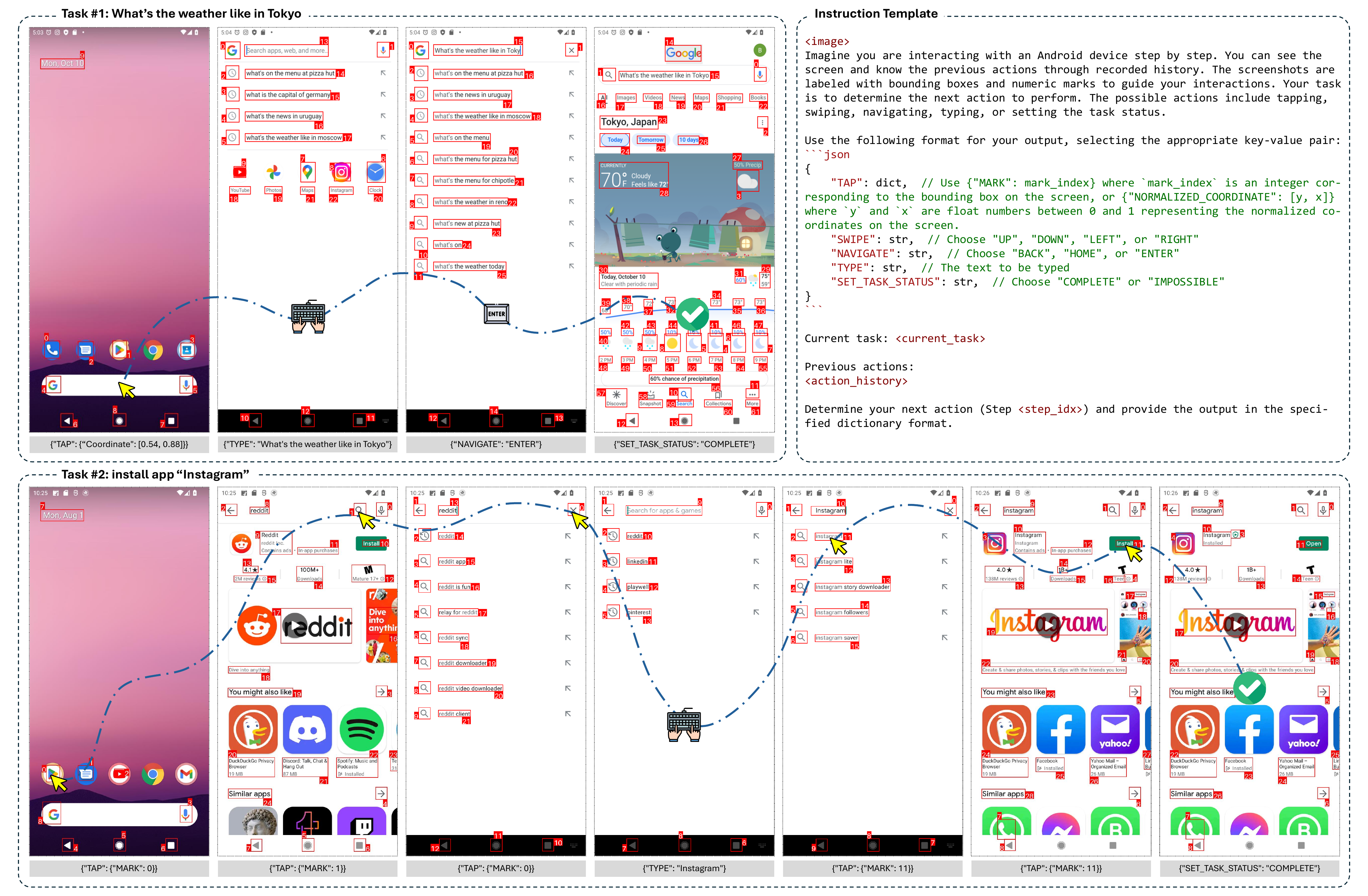}
    \caption{\textbf{Examples for mobile UI navigation sample}. We prompt the model with two tasks: ``What's the weather like in Tokyo" and ``Install app `Instagram'". The model take actions sequentially given the new observation and history action information.}
    \label{fig:supp_aitw_ui}
\end{figure*}

Given the performant UI navigation performance across different tasks, we show some Mobile UI navigation samples in Fig.~\ref{fig:supp_aitw_ui}. We prompt the model to complete two daily tasks starting from the home page: ``What's the weather like in Tokyo" and ``Install app `Instagram'". Despite that our model is never trained with the full trajectory, it can handle the tasks in the wild pretty well.

\subsection{Robotics Manipulation}

\renewcommand{\thesubfigure}{\alph{subfigure}} 

\renewcommand{\thesubfigure}{\alph{subfigure}} 
\begin{figure*}[!t]
\begin{subfigure}{1.0\textwidth}
    \includegraphics[width=1.0\linewidth]{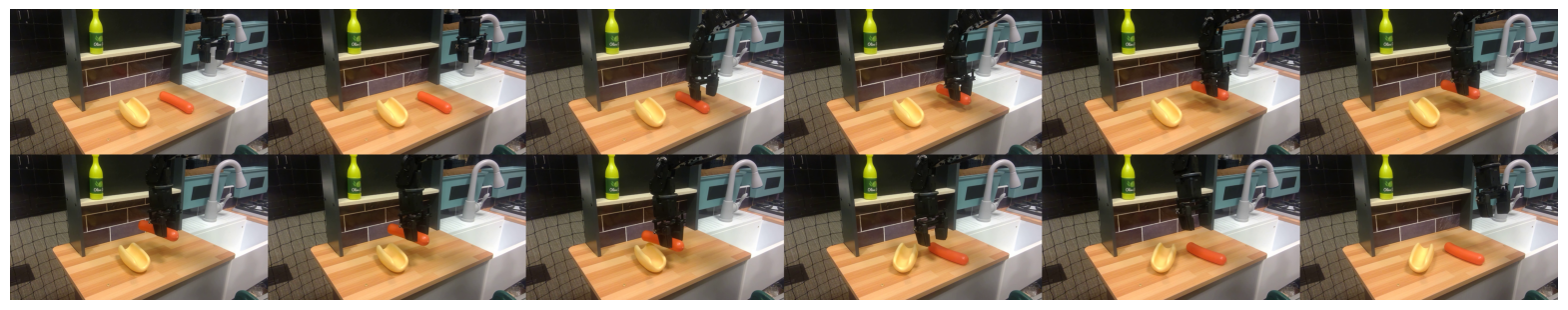}      
    \caption{Robot policy rollout for task ``Put the sausage to hotdog'' for OpenVLA model. (\textcolor{red}{Failure})}
\end{subfigure}
\begin{subfigure}{1.0\textwidth}
    \includegraphics[width=1.0\linewidth]{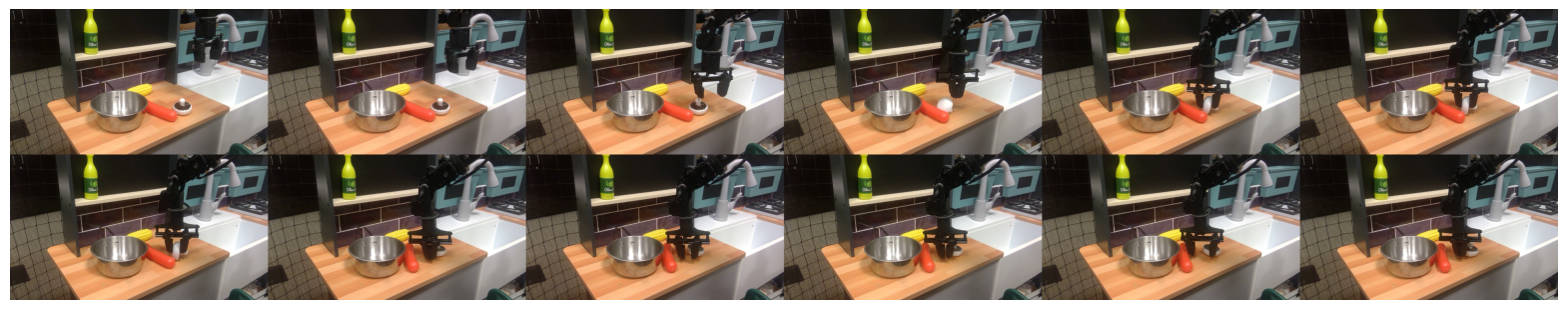}      
    \caption{Robot policy rollout for task ``Pick up the mushroom to the pot'' for OpenVLA model. (\textcolor{red}{Failure})}
\end{subfigure}

\begin{subfigure}{1.0\textwidth}
    \includegraphics[width=1.0\linewidth]{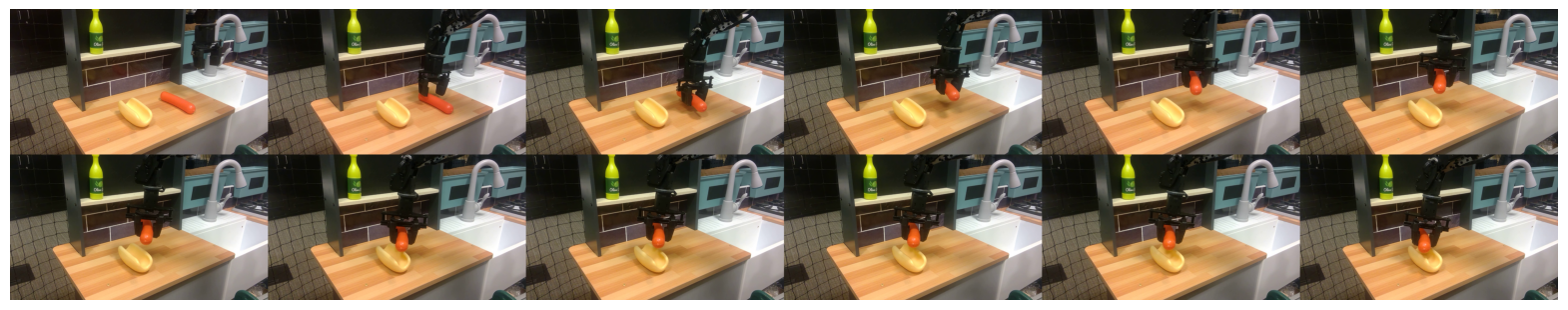}   
    \caption{Robot policy rollout for task ``Put the sausage to hotdog'' for Magma model. (\textcolor{lightgreen}{Success})}    
\end{subfigure}
\begin{subfigure}{1.0\textwidth}
    \includegraphics[width=1.0\linewidth]{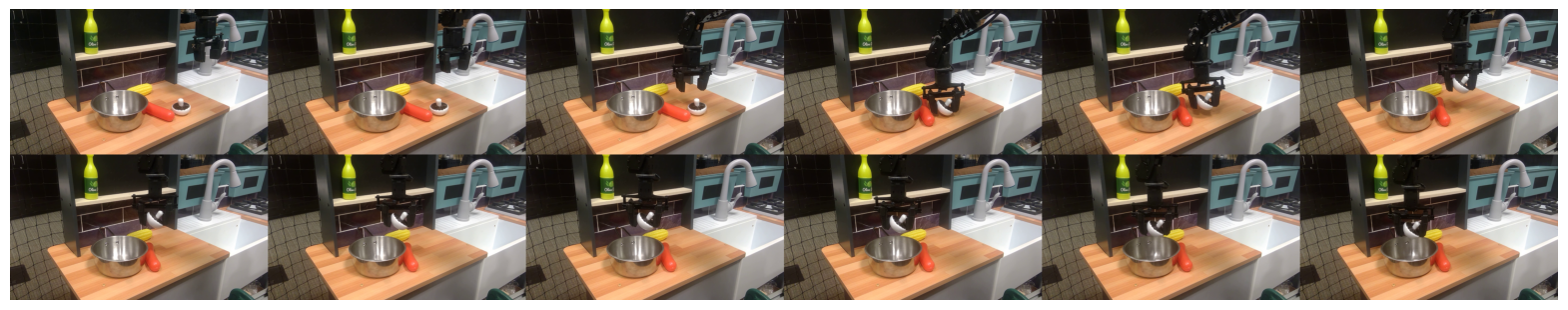}   
    \caption{Robot policy rollout for task ``Pick up the mushroom to the pot'' for Magma model. (\textcolor{lightgreen}{Success})}    
\end{subfigure}

    \caption{\textbf{Comparison between OpenVLA (top two rows) and Magma (bottom two rows) for real robot manipulation task.} The two robot policies starts with the same initial stage and asked to perform exactly the same task. The whole task requires precise spatial understanding and planning for the model. For both tasks, OpenVLA failed to accomplish while our model successfully handle.}
    \label{fig:supp_real_robot}
\end{figure*}

We further show the real robot manipulation rollout for OpenVLA and Magma model. As discussed in our main paper, our model exhibits much better generalization ability to different real robot manipulation tasks. In Fig.~\ref{fig:supp_real_robot}, we qualitatively show how two models handle a complicated task of ``Pick up the sausage and put it inside the hotdog''. Thanks to the proposed pretraining techniques, our Magma model can not only precisely pick up the sausage but also move smoothly to the top of the hotdog, demonstrating superior spatial understanding and reasoning capability compared with the counterpart.


\end{document}